\def\cvprPaperID{10803}
\def\confName{CVPR}
\def\confYear{2023}

\def\authorBlock{
    Kihong Kim\thanks{Equal contribution} \textsuperscript{\rm ,1}   \qquad
    Yunho Kim\samethanks{} \textsuperscript{\rm ,1} \qquad
    Seokju Cho\textsuperscript{\rm 2} \qquad
    Junyoung Seo\textsuperscript{\rm 2} \\
    Jisu Nam\textsuperscript{\rm 2} \qquad
    Kychul Lee\textsuperscript{\rm 1} \qquad
    Seungryong Kim\thanks{Corresponding author} \textsuperscript{\rm ,2} \qquad
    Kwang Hee Lee\samethanks{} \textsuperscript{\rm ,1} \vspace{1em} \\ 
    \textsuperscript{\rm 1}VIVE STUDIOS \qquad \textsuperscript{\rm 2}Korea University \\
    {\tt\small \textsuperscript{\rm 1}\{hxngiee,youknowyunho,lkc880425,lucas\}@vivestudios.com} \\
    {\tt\small \textsuperscript{\rm 2}\{seokju\_cho,se780,jisu\_nam,seungryong\_kim\}@korea.ac.kr}
}

\newcommand*\samethanks[1][\value{footnote}]{\footnotemark[#1]}

\newif\ifreview 
\newif\ifarxiv \newcommand{\arxiv}{\arxivtrue}
\newif\ifcamera 
\newif\ifrebuttal 

\arxiv 

\pdfoutput=1
\documentclass[10pt,twocolumn,letterpaper]{article}

\usepackage{algorithm, algorithmic}
\usepackage{xspace}
\usepackage{mathtools,amssymb}
\usepackage{textcomp}

\newcommand{\tabPerformance}{
\begin{table}[t]
    \centering
\resizebox{\linewidth}{!}{
{\small
\begin{tabular}{l|rccc|ccc}
\toprule

\textbf{Model} & \textbf{Arc↑} & \textbf{Arc-R↑} & \textbf{Cos↑} & \textbf{Cos-R↑} & \textbf{Expr↓} & \textbf{Pose↓} &  \textbf{Shp↓} \\

\midrule\midrule
SimSwap\cite{chen2020simswap}        & $\dagger$  & $\dagger$  & \underline{0.597} & 0.756  & \textbf{0.033} & \textbf{0.0005} & \underline{0.0256}  \tabularnewline
HifiFace\cite{wang2021hififace}      & \underline{0.575}  & \underline{0.816}  & 0.565 & 0.792  & 0.048 & 0.0007 & 0.0299 \tabularnewline
InfoSwap\cite{gao2021information}    & $\dagger$ & $\dagger$ & 0.570 & \textbf{0.841} & 0.052 & 0.0010 & 0.0360 \tabularnewline
MegaFS\cite{zhu2021one}              & $\dagger$  & $\dagger$  & 0.343 & 0.553  & 0.046 & 0.0024 & 0.0299 \tabularnewline
FaceShifter\cite{li2019faceshifter}  & $\dagger$  & $\dagger$  & 0.534   & 0.657.  & 0.061  & 0.0013 & \textbf{0.0235} \tabularnewline
DeepFakes\cite{DeepFakesHttpsGithub2021}    & 0.443  & 0.686  & 0.437   & 0.635   & 0.078  & 0.0022 & 0.0314 \tabularnewline

\midrule
(Cos) DiffFace($\hat{T}=40$) & \textbf{0.620}  & \textbf{0.859} & $\dagger$ & $\dagger$  & 0.044 & 0.0009 & 0.0269 \tabularnewline
(Arc) DiffFace($\hat{T}=40$) & $\dagger$ & $\dagger$ & \textbf{0.602} & \underline{0.816} & \underline{0.043} & \underline{0.0008} & 0.0283 \tabularnewline
\bottomrule
\end{tabular}
}} 

\caption{\textbf{Quantitative comparison on FaceForensics++~\cite{rosslerFaceForensicsLearningDetect2019} dataset.} 
 (Arc), (Cos) denotes that model was trained using ArcFace, CosFace respectively. $\dagger$ denotes that model cannot be evaluated because they are trained and evaluated using same identity embedder. Expr, Pose and Shp denotes expression, pose, and shape distance obtained by 3D face model \cite{RingNet:CVPR:2019} respectively. Please refer to Sec.~\ref{sec:experiment} for more details.}
\label{tab:comparison}\vspace{-10pt}
\end{table}
}

\newcommand{\tabAblation}{
\begin{table}[t]
    \centering
\resizebox{\linewidth}{!}{
{\small
\begin{tabular}{ll|rccc|ccc}
\toprule

&\textbf{Model} & \textbf{Arc↑} & \textbf{Arc-R↑} & \textbf{Cos↑} & \textbf{Cos-R↑} & \textbf{Expr↓} & \textbf{Pose↓} &  \textbf{Shp↓} \\

\midrule\midrule
\textbf{(I)} & Unconditional & 0.486 & 0.803 & 0.455 & 0.801 & 0.063 & 0.0016 & 0.0358 \tabularnewline
\textbf{(II)} & Conditional \textit{+} ($\lambda_{id}=0$) & 0.455 & 0.751 & 0.535 & 0.781 & \underline{0.040} & \underline{0.0008} & \textbf{0.0220} \tabularnewline
\midrule
\textbf{(III)} &(Cos) $\hat{T}=30$  & 0.598           & 0.813  & $\dagger$ & $\dagger$ & 0.037 & 0.0007 & \underline{0.0226}\tabularnewline
\textbf{(IV)} &(Cos) $\hat{T}=40$  & \underline{0.620}           & \underline{0.859} & $\dagger$ & $\dagger$  & 0.044 & 0.0009 & 0.0269 \tabularnewline

\textbf{(V)} &(Cos) $\hat{T}=50$  & \textbf{0.634}  & \textbf{0.888}  & $\dagger$ & $\dagger$ & 0.050 & 0.0011 & 0.0303\tabularnewline
\midrule
\textbf{(VI)} &(Arc) $\hat{T}=30$  & $\dagger$       & $\dagger$  & 0.580 & 0.766 & \textbf{0.035} & \textbf{0.0006} & 0.0240\tabularnewline

\textbf{(VII)} &(Arc) $\hat{T}=40$  & $\dagger$       & $\dagger$  & \underline{0.602} & \textbf{0.816} & 0.043 & \underline{0.0008} & 0.0283 \tabularnewline
\textbf{(VIII)} &(Arc) $\hat{T}=50$  & $\dagger$       & $\dagger$  & \textbf{0.603} & \textbf{0.816} & 0.049 & 0.0009 & 0.0311\tabularnewline
\bottomrule
\end{tabular}
}} 
\caption{\textbf{Quantitative ablation study.} Performance of various configurations on FF++. $\lambda_{id}=0$ denotes that identity guidance was not used, and $\hat{T}$ denotes the timestep where mask intensity becomes one. For other notations, please refer the caption of Table~\ref{tab:comparison}.
}
\label{tab:synth}\vspace{-10pt}
\end{table}
}

\ifreview \usepackage[review]{cvpr} \fi
\ifarxiv \usepackage[pagenumbers]{cvpr} \fi
\ifrebuttal \usepackage[rebuttal]{cvpr} \fi
\ifcamera \usepackage{cvpr} \fi

\usepackage{graphicx}
\usepackage{amsmath}
\usepackage{amssymb}
\usepackage{booktabs}

\usepackage{times}
\usepackage{microtype}
\usepackage{epsfig}
\usepackage[table,xcdraw]{xcolor}
\usepackage{caption}
\usepackage{float}
\usepackage{placeins}
\usepackage{color, colortbl}
\usepackage{stfloats}
\usepackage{enumitem}
\usepackage{tabularx}
\usepackage{xstring}
\usepackage{multirow}
\usepackage{xspace}
\usepackage{url}
\usepackage{subcaption}
\usepackage{xcolor}
\usepackage[hang,flushmargin]{footmisc}

\ifcamera \usepackage[accsupp]{axessibility} \fi

% ...

\ifarxiv  \fi

\newcommand{\R}[1]{{%
    \textbf{%
        \ifstrequal{#1}{1}{\textcolor{red}{R#1}}{%
        \ifstrequal{#1}{2}{\textcolor{blue}{R#1}}{%
        \ifstrequal{#1}{3}{\textcolor{magenta}{R#1}}{%
        \ifstrequal{#1}{4}{\textcolor{teal}{R#1}}{%
                           \textcolor{cyan}{R#1}%
        }}}}%
    }%
}}

\usepackage{xr-hyper}

\makeatletter
\newcommand*{\addFileDependency}[1]{
  \typeout{(#1)}
  \@addtofilelist{#1}
  \IfFileExists{#1}{}{\typeout{No file #1.}}
}

\makeatother

\usepackage[pagebackref,breaklinks,colorlinks]{hyperref}
\usepackage[capitalize]{cleveref}
\crefname{section}{Sec.}{Secs.}
\crefname{table}{Table}{Tables}
\crefname{figure}{Fig.}{Figs.}

\begin{document}
\title{DiffFace: Diffusion-based Face Swapping with Facial Guidance}
\author{\authorBlock}

\twocolumn[{
\maketitle
\thispagestyle{empty}
\begin{figure}[H]
    \vspace{-20pt}
    \hsize=1.0\textwidth
    \begin{center}
    \includegraphics[width=0.9\textwidth]{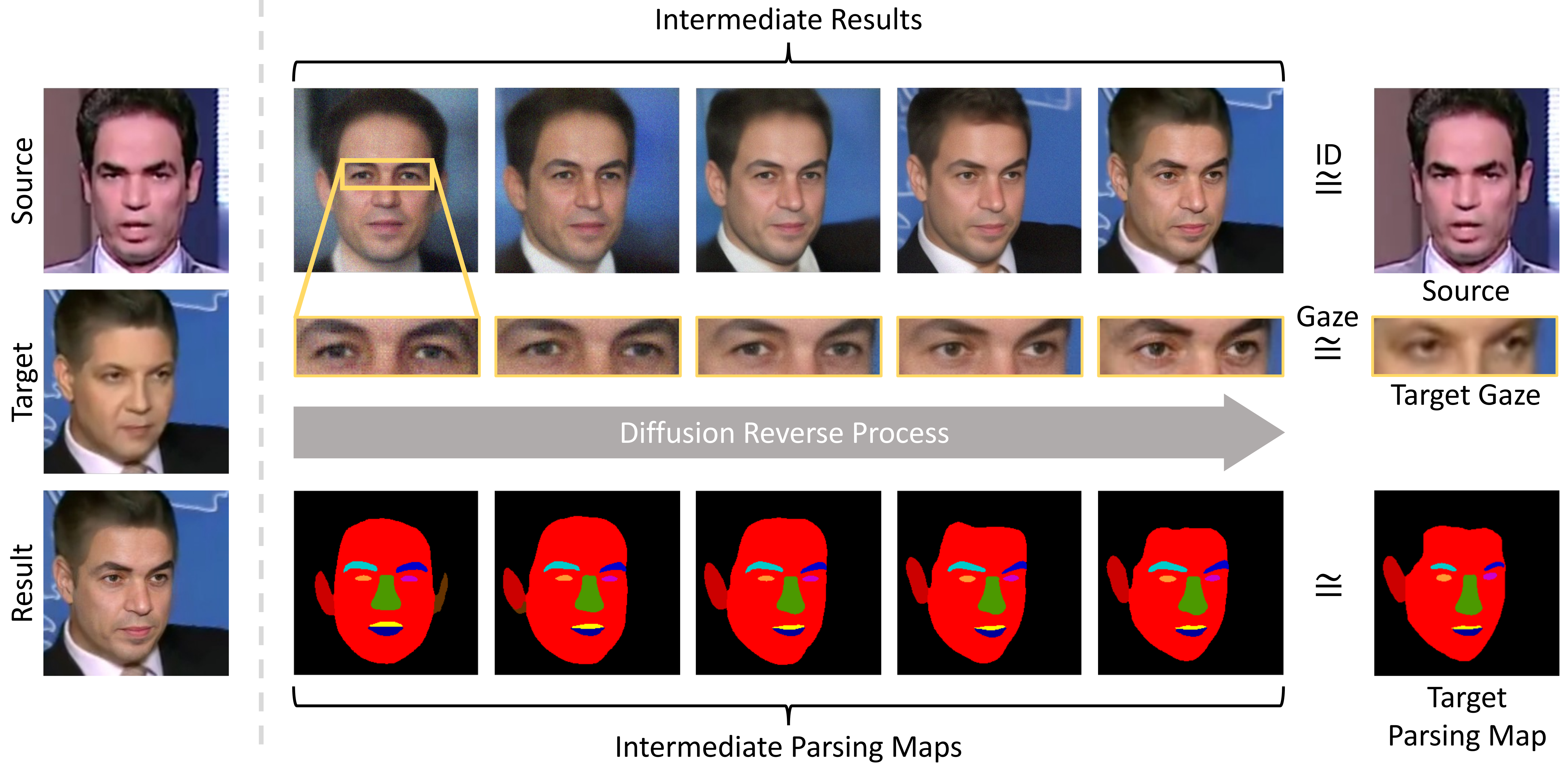}
    \vspace{-10pt}
    \caption{\textbf{Visualization of our novel diffusion-based face swapping framework, called DiffFace.} DiffFace gradually produces images with source identity and target attributes such as gaze, structure and pose.}
    \label{fig:main}
    \end{center}
    \end{figure}
}]

{
  \renewcommand{\thefootnote}%
    {\fnsymbol{footnote}}
  \footnotetext[1]{Authors contributed equally.}
  \footnotetext[2]{Corresponding author.}
}
\begin{abstract}
In this paper, we propose a novel diffusion-based face swapping framework, called DiffFace, composed of training ID Conditional DDPM, sampling with facial guidance, and a target-preserving blending.\footnote{Project Page : \url{https://hxngiee.github.io/DiffFace/}} In specific, in the training process, the ID Conditional DDPM is trained to generate face images with the desired identity. In the sampling process, we use the off-the-shelf facial expert models to make the model transfer source identity while preserving target attributes faithfully. During this process, to preserve the background of the target image, we additionally propose a target-preserving blending strategy. It helps our model to keep the attributes of the target face from noise while transferring the source facial identity. In addition, without any re-training, our model can flexibly apply additional facial guidance and adaptively control the ID-attributes trade-off to achieve the desired results.
To the best of our knowledge, this is the first approach that applies the diffusion model in face swapping task. Compared with previous GAN-based approaches, by taking advantage of the diffusion model for the face swapping task, DiffFace achieves better benefits such as training stability, high fidelity, and controllability. Extensive experiments show that our DiffFace is comparable or superior to the state-of-the-art methods on the standard face swapping benchmark. 
\end{abstract}


\vspace{-15pt}
\section{Introduction}
\label{sec:intro}
Generative adversarial networks (GANs)~\cite{goodfellow2014generative} have shown incredible empirical success in image generation~\cite{zhu2016generative, DBLP:journals/corr/MirzaO14, brock2018large, https://doi.org/10.48550/arxiv.1710.10196, karras2019style,pix2pix2017} tasks. Recently, most of face swapping tasks~\cite{chen2020simswap, wang2021hififace, DeepFakesHttpsGithub2021, gao2021information, li2019faceshifter, zhu2021one} have been proposed based on GANs. Despite their empirical successes, it has been well known that the training of GANs is inherently unstable\cite{https://doi.org/10.48550/arxiv.1802.05957} due to the min-max optimization problem of the generator and discriminator. To alleviate this problem, complex architecture, various loss functions, and extensive hyperparameter tuning have been required. In addition, tuning hyperparameters becomes even more challenging if external models are combined.


Face swap is a task to synthesize an image with the identity of the source image while preserving the target image's attributes (e.g., expression, pose, and shape). Off-the-shelf expert models trained on specific purposes can be used in the face swap task. For instance, ID embedder can be utilized to constrain the synthesized image's identity to follow the source's identity\cite{chen2020simswap, li2019faceshifter}. Nevertheless, balancing identity and attributes is  one of the most challenging problems. GAN-based face swapping tasks~\cite{chen2020simswap, wang2021hififace} maintain this balance through a combination of loss functions related with ID and facial attributes. However, it is necessary to perform multiple training to find the desired hyperparameters. Due to the trade-off between ID and attributes, a greater focus on maintaining natural attributes often makes it difficult to achieve satisfactory results in transferring the source face ID to a synthesized face. 

Recently, diffusion models~\cite{dhariwal2021diffusion, nichol2021improved, preechakul2022diffusion, rombach2022high} have attracted much attention as an alternative to GANs~\cite{goodfellow2014generative}. Diffusion models, as opposed to GANs, enable more stable training ,showing desirable results in terms of diversity and fidelity. To tackle the trade-off between fidelity and diversity, classifier guidance~\cite{dhariwal2021diffusion} is introduced to guide the diffusion model. Such a guidance technique is also widely used in conditional generation~\cite{liu2021more,zhao2022egsde}, especially in text-to-image generation~\cite{nichol2021glide,ramesh2022hierarchical, saharia2022photorealistic,avrahami2022blended}. Despite various advantages of the guidance technique, we stress the usability of the external module as guidance at test time.
In this paper, we propose a novel diffusion-based face swap framework, named \textit{DiffFace}, which is composed of training ID Conditional DDPM, sampling with facial guidance, and a target-preserving blending strategy. First of all, we present an ID Conditional DDPM. In the training process, the ID Conditional DDPM is trained to generate face images with the desired identity. We make the diffusion model aware of facial identity by not only injecting the identity feature vector from the ID embedder but also posing additional constraint with the identity similarity loss. In the sampling process, we use facial guidance driven by various pretrained experts to enable the model to transfer source identity while preserving target attributes faithfully as illustrated in Fig.~\ref{fig:main}. During this process, to preserve the background of the target image and obtain the desired face swapping result, we additionally propose a target-preserving blending strategy. By gradually increasing the facial mask intensity over the time of the diffusion process, it prevents our model from completely forgetting the attributes of the target face by noise while transferring the source facial identity. In addition, our model can flexibly apply various facial guidance and adaptively control the ID-attributes trade-off to achieve the desired results without any re-training. Compared with the GAN-based face swapping works, \textit{DiffFace} achieves better benefits such as training stability, high fidelity, and controllability. Experiments demonstrate that our method surpasses other state-of-the-art methods on standard face swapping benchmark. 
\section{Related Work}
\label{sec:related}

\subsection{Diffusion Model}
Diffusion models~\cite{ho2020denoising, nichol2021improved} have attained much attention as a generative model showing desirable qualities while maintaining a higher distribution coverage. The sampling process of the diffusion models can be intuitively viewed as a denoising process. Utilizing the characteristics of the gradual denoising process, various works~\cite{meng2021sdedit, lugmayr2022repaint, song2020score, saharia2022palette, seo2022midms} have achieved remarkable results in the field of conditional generation~\cite{choi2021ilvr}, local image editing~\cite{meng2021sdedit, lugmayr2022repaint, song2020score}, and image translation~\cite{meng2021sdedit, saharia2022palette, song2020score, seo2022midms}. On the one hand, ADM~\cite{dhariwal2021diffusion} proposed to interpret the gradient of the external classifier as guidance and inject it during the reverse process. This not only improved the sample quality but also succeeded in conditional generation. These guidance techniques have been widely used, including text-to-image generation~\cite{nichol2021glide, ramesh2022hierarchical, saharia2022photorealistic, rombach2022high}. Also, recent work~\cite{avrahami2022blended} enabled text-guided editing of local area by injecting CLIP guidance~\cite{radford2021learning} only into the local area. Although various methods have been proposed, no research directly tackled the face swapping task. For the first time, we propose a high-fidelity and controllable face swapping framework based on the conditional diffusion model.

\subsection{Face Swap Model}
\paragraph{Structural Prior-Guided Models.} Previous face swapping methods such as HifiFace~\cite{wang2021hififace} combine structural information extracted from 3D Morphable Model~\cite{blanzMorphableModelSynthesis1999} with GANs to produce 3D shape-aware identity. With improving the geometric structure of generated images, FSGAN~\cite{nirkin2019fsgan} attempts to reenact the source image to match the target image using facial key points and segmentation. However, inaccurate structural or facial key points information makes previous models struggle to generate high-fidelity results. \vspace{-10pt}

\paragraph{Reconstruction-Based Models.}  DeepFakes~\cite{DeepFakesHttpsGithub2021} was trained to swap faces between paired identities. However, this method can only be applied to one specific identity. Subject-agnostic face swap methods such as Faceshifter~\cite{liFaceShifterHighFidelity2019} and SimSwap~\cite{chen2020simswap} were proposed to overcome the limitation of identity-specific face swap. These subject-agnostic models typically tune intermediate features of the target image to inject the identity of the source image. Even though these methods could synthesize arbitrary identities, they are insufficient to generate high fidelity results. 

\vspace{-10pt}
\paragraph{StyleGAN-Based Models.} Recently, StyleGAN~\cite{karras2019style, karras2020analyzing} based face swap model~\cite{xu2022styleswap} has emerged as a solution for high resolution face swap. These models relied on StyleGAN architectures \cite{karras2019style,karras2020analyzing,karras2021alias}, which have powerful latent space representation and expressibility. MegaFS \cite{zhu2021one} proposed to invert source and target image into latent space, then fuses these two features appropriately and feeds it into StyleGAN generator to obtain the results. Other work \cite{xu2022high} focused more on separating identity and pose information. InfoSwap \cite{gao2021information} proposed identity contrastive loss that better disentangles StyleGAN latent space. The StyleGAN-based face swap methods~\cite{xu2022styleswap, zhu2021one, xu2022high, gao2021information} fusing the source and target's multi-resolution features typically failed to synthesize fine details.

\section{Preliminaries: Denoising Diffusion Probabilistic Models}

\begin{figure}[]
    \begin{center}
    \includegraphics[width=1.0\linewidth]{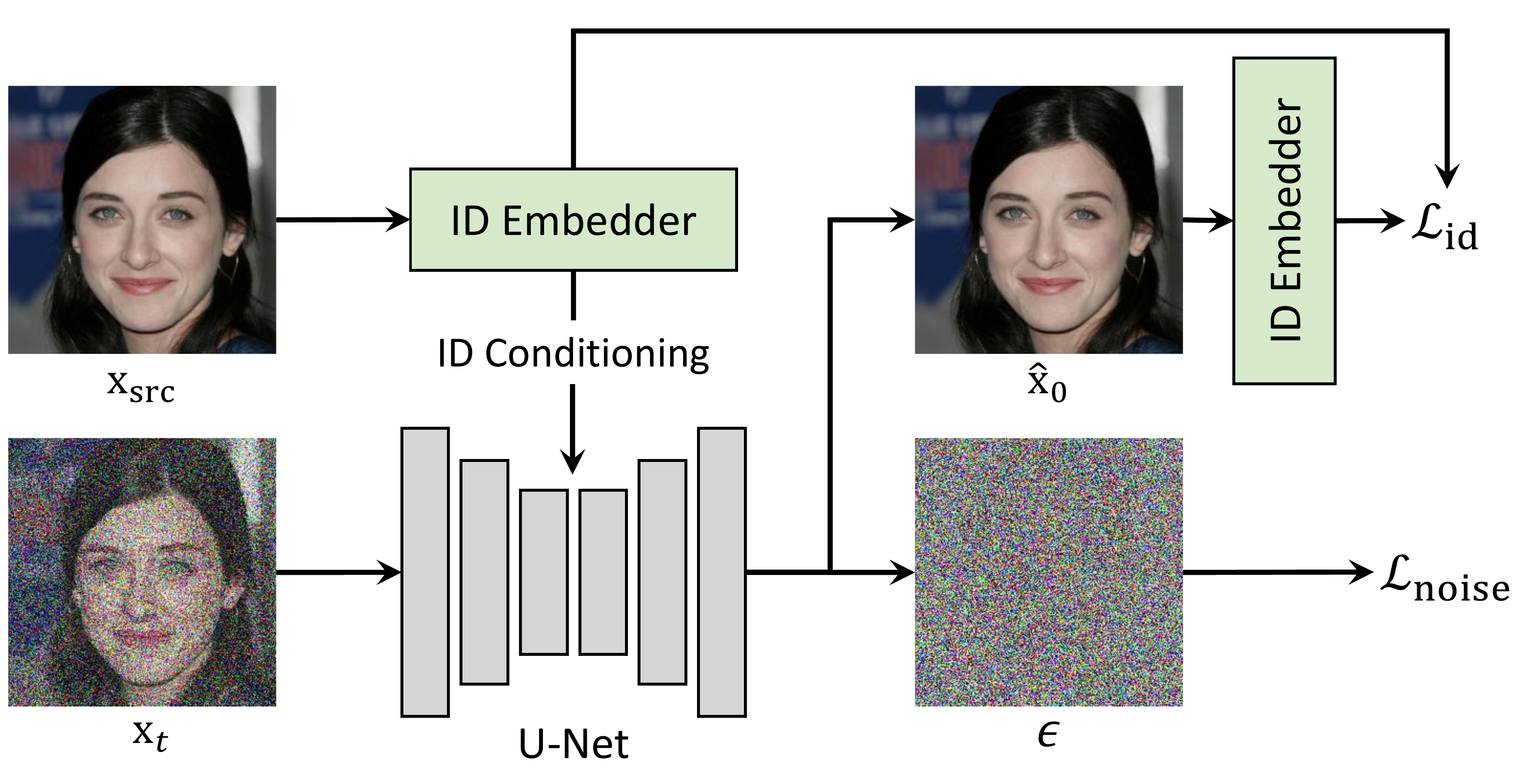}
    \vspace{-10pt}
    \caption{\textbf{Training procedure for ID Conditional DDPM.} $\widehat{\mathbf{x}}_0$ denotes the predicted denoised image at timestep t. In each denoising step, we condition ID vector of the source image into the U-Net. Then, we compute cosine similarity loss between the ID vectors of source and $\widehat{\mathbf{x}}_0$. }
    \label{fig:method_idcond}\vspace{-10pt}
    \end{center}
\end{figure}

Diffusion models generate a realistic image from a standard Gaussian distribution by reversing a recurrent noising process~\cite{ho2020denoising}. The forward process gradually alters to Gaussian distribution from the data $\mathbf{x}_0 \sim q(\mathbf{x}_0)$, which is defined as $q(\mathbf{x}_t|\mathbf{x}_{t-1}):=\mathcal{N}(\mathbf{x}_{t};\sqrt{1-\beta_t}\mathbf{x}_{t-1}, \beta_t \mathbf{I})$, where $\beta_t$ is a predefined variance schedule. 
In addition, the reverse process is as follows:
\begin{equation} \label{eq:forward_ddpm}
    p_\theta(\mathbf{x}_{t-1}|\mathbf{x}_t):=\mathcal{N}(\mathbf{x}_{t-1}; \mu_\theta (\mathbf{x}_t, t), \sigma_\theta(\mathbf{x}_t, t)\mathbf{I}),
\end{equation} 
where $\mu_\theta(\cdot)$ and $\sigma_\theta(\cdot)$ can be parameterized using deep neural networks. But, in practice, it is known that using noise approximation model $\epsilon_\theta(\mathbf{x}_t, t)$ worked best instead of using $\mu_\theta (\mathbf{x}_t, t)$.~\cite{ho2020denoising}
Thus, $\mu_\theta (\mathbf{x}_t, t)$ can be induced as:
\begin{equation}
    \mu_\theta(\mathbf{x}_t, t)= \frac{1}{\sqrt{1-\beta_t}}\left(\mathbf{x}_t - \frac{\beta_t}{\sqrt{1-\alpha_t}} \epsilon_\theta(\mathbf{x}_t, t)\right).
    \label{eq:reverse_ddpm}
\end{equation}
Given $\mathbf{x}_t$, the reverse process of the diffusion model usually outputs $\mathbf{x}_{t-1}$. But, we can also directly derive $\hat{\mathbf{x}}_0$ which is the fully denoised prediction given $\mathbf{x}_t$ by $\hat{\mathbf{x}}_0=f_\theta(\mathbf{x}_{t}, t)$ where
\begin{equation}\label{eq:direct_pred}
  f_\theta(\mathbf{x}_{t}, t):= \frac{\mathbf{x}_{t} - \sqrt{1-\alpha_t}\epsilon_{\theta}(\mathbf{x}_t,t)}{\sqrt{\alpha_{t}}}.
 \end{equation}
We utilize this fully denoised prediction $\hat{\mathbf{x}}_0$ for facial expert modules.

Meanwhile, ADM~\cite{ramesh2022hierarchical} proposes a guidance technique for the diffusion model. ADM~\cite{ramesh2022hierarchical} trains a classifier $p(y|\mathbf{x}_t,t)$, which takes a noised image as input, and regards the gradient of the classifier as a guidance for the diffusion models:
\begin{align}
    p_\theta(\mathbf{x}_{t-1}|\mathbf{x}_t,y):=
    \mathcal{N}(\mu+s\nabla_{\mathbf{x}_t}p(y|\mathbf{x}_t,t), \sigma\mathbf{I}),
    \label{eq:classifier_guidance}
\end{align}
where $s$ is a constant for the guidance scale and $\mu$, $\sigma$ are $\mu_\theta(\mathbf{x}_t, t)$ and $\sigma_\theta(\mathbf{x}_t, t)$, respectively. Note that the guidance technique proposed by ADM~\cite{ramesh2022hierarchical} uses an unconditional diffusion model, which is different in that our work utilizes ID Conditional DDPM. Although the diffusion model has the advantage of adding various guidance techniques while guaranteeing strong generation capabilities, employing the diffusion model to face swapping has not yet been explored due to the difficulties described below.
\label{sec:related}
\section{Methodology}
\label{sec:method}

\newcommand{\bepsilon}{{\boldsymbol{\epsilon}}}

\begin{algorithm}[t]
    \footnotesize
    \caption{Training ID Conditional DDPM}
    \begin{algorithmic}

        \REPEAT
            \STATE $\mathbf{x}_0 \sim q(\mathbf{x}_0)$ 
            \STATE $t \sim \mathrm{Uniform}(\{1, \dotsc, T\})$
            \STATE ${\bepsilon} \sim \mathcal{N}(\mathbf{0}, \mathbf{I})$

            \STATE $\mathbf{x}_t \sim \mathcal{N}(\sqrt{\bar{\alpha}_t} \mathbf{x}_0, (1-\bar{\alpha}_t) \mathbf{I})$
            \STATE  ${\mathbf{v}}_{\mathrm{id}} \gets \mathcal{D}_{\textit{I}}({\mathbf{x}}_0)$
            
            \STATE $\widehat{\mathbf{x}}_0 \gets \frac{\mathbf{x}_t - \sqrt{1-\bar{\alpha}_t} \epsilon_{\theta}(\mathbf{x}_t, t,\mathbf{v}_{\mathrm{id}})}{\sqrt{\bar{\alpha}_t}} $ 
            \STATE $\widehat{\mathbf{v}}_{\mathrm{id}}  \gets \mathcal{D}_{\mathrm{I}}(\widehat{\mathbf{x}}_0)$

            \STATE Take gradient descent step on
             \STATE $\nabla_{\theta} \big(\left\| \bepsilon - \bepsilon_\theta(\mathbf{x}_t, t, \mathbf{v}_{\mathrm{id}}) \right\|^{2}_2 $+$ \lambda  \left\| \mathbf{v}_{\mathrm{id}} - \widehat{\mathrm{\mathbf{v}}}_{\mathrm{id}} \right\|^2_2\big)$
            
        \UNTIL{converged}
        
    \end{algorithmic}
    \label{alg:id_conditional_guided_diffusion}
\end{algorithm}

In this section, we describe our method called \textit{DiffFace}. The whole process is divided into training ID Conditional DDPM, sampling with facial guidance, and target-preserving blending. We propose ID Conditional DDPM which constructs diffusion model suitable for face swap tasks. Then, we describe the facial guidance designed to synthesize the desired image in the diffusion process. Lastly, we introduce target-preserving blending, which can help ID Conditional DDPM preserve the target facial detail better.

Throughout this paper, we use $\mathbf{x}_{\mathrm{src}}$, $\mathbf{x}_{\mathrm{targ}}$, and $\mathbf{x}_{\mathrm{swap}}$ to represent the source, target, and synthesized image, respectively. 
Also, let us denote $\mathcal{D}$ by the pretrained guidance network. Specifically,  $\mathcal{D}_{\textit{I}}$ denotes identity embedder \cite{wang2018cosface, deng2019arcface},  $\mathcal{D}_{\textit{F}}$ denotes face parser \cite{yubilateral}, and $\mathcal{D}_{\textit{G}}$ denotes gaze estimator \cite{Park2018ETRA}.  

\subsection{ID Conditional DDPM}
Our key idea for face swapping with the diffusion model is to inject the identity feature into the diffusion model. While previous methods have studied these conditioning problems~\cite{rombach2022high, saharia2022palette,ramesh2022hierarchical, saharia2022photorealistic} extensively, no study infuses identity information as a condition into the diffusion model. Thus, we employ the structure of the conditional diffusion model, where additional information can be injected. As shown in Fig.~\ref{fig:method_idcond} we first inject the source image ${\mathbf{x}}_{\mathrm{src}}$ in the identity embedder $\mathcal{D}_{\textit{I}}$ (e.g., ArcFace~\cite{deng2019arcface} and CosFace~\cite{wang2018cosface}) to obtain the source identity ${\mathbf{v}}_{\mathrm{id}}$:
\begin{equation}
    {\mathbf{v}}_{\mathrm{id}} = \mathcal{D}_{\textit{I}}({\mathbf{x}}_{\mathrm{src}}),
    \label{eq:identity_embed}
\end{equation} 
Then, we embed source identity $\mathbf{v}_\mathrm{id}$ into the diffusion model $\epsilon_\theta(\mathbf{x}_t,t,\mathbf{v}_\mathrm{id})$, where $\mathbf{x}_t$ is a noisy version of source image $\mathbf{x}_\mathrm{src}$ at timestep $t$ using Eq.~\ref{eq:forward_ddpm}.
\begin{figure*}[t]
    \begin{center}
    \begin{subfigure}{1.0\textwidth}
        \centering
        \includegraphics[width=\textwidth]{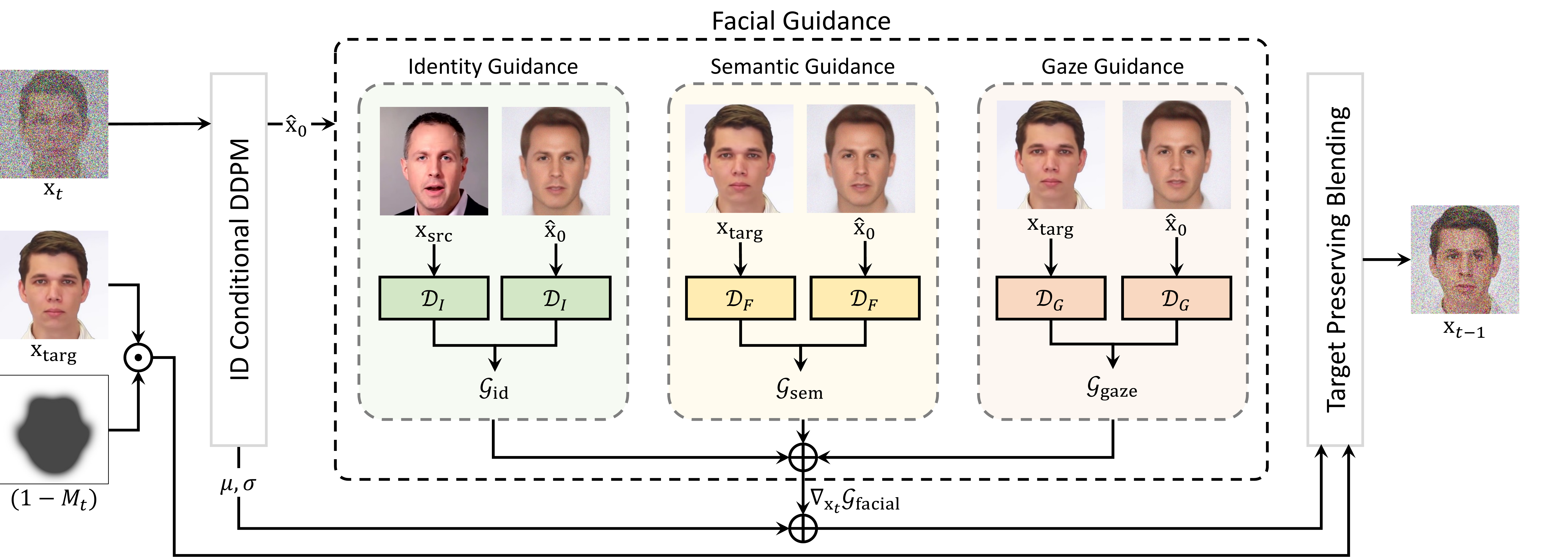}
    \end{subfigure}
    \end{center}
    \vspace{-10pt}
    \caption{\label{fig:arch} \textbf{Sampling procedure with facial guidance.} Given facial mask and target image, we leverage the generative process of diffusion model to synthesize coherent face swapping results. Facial guidance induces the masked region to have the desired facial attributes. Our facial guidance module is composed of three components, which can be found in the bottom of the figure.}
\end{figure*}

\vspace{-10pt}
\paragraph{Loss Functions.}
Our DiffFace learns a reverse process that reverts the forward process (Fig.~\ref{fig:method_idcond}). Given a noisy source image $\mathbf{x}_{t}$, our ID Conditional DDPM aims to predict the noise preserving the source identity $\mathbf{v}_\mathrm{id}$ by using the denoising score matching loss:
\begin{equation}
  \mathcal{L}_{\mathrm{noise}} = ||\epsilon - \epsilon_\theta(\mathbf{x}_t, t, \mathbf{v}_{\mathrm{id}})||^2_2 ,
  \label{eq:training_dsmloss}
\end{equation}
where $\epsilon$ is a noise added to $\mathbf{x}_t$. 
At the same time, we propose an identity loss for the diffusion model to preserve the facial identity effectively. Since expert models~\cite{deng2019arcface, wang2018cosface, yubilateral, Park2018ETRA} are trained on clean images, 
we estimate the fully denoised image $\widehat{\mathbf{x}}_0$ from each transition $\mathbf{x}_{t}$ during the denoising diffusion process:
\begin{equation}
    \widehat{\mathbf{x}}_0 = f_\theta(\mathbf{x}_t,t,\mathbf{v}_{\mathrm{id}}),
\end{equation}
where $f_\theta(\cdot)$ is a function to estimate a fully denoised image, which is defined in Eq.~\ref{eq:direct_pred}.
Concretely, we induce the identity of predicted denoised image $\widehat{\mathbf{v}}_{\mathrm{id}}$:
\begin{equation}
    \widehat{\mathbf{v}}_{\mathrm{id}} = \mathcal{D}_{\textit{I}}(\widehat{\mathbf{x}}_{0}), 
\end{equation}
to be equal to the source identity $\mathbf{v}_{\mathrm{id}}$ at every timestep $t$. Using this predicted identity, the proposed identity loss is as follows:
\begin{equation}
   \mathcal{L}_\mathrm{id} = 1 - \cos(\mathbf{v}_{\mathrm{id}}, \widehat{\mathbf{v}}_{\mathrm{id}}),
   \label{eq:training_idloss}
 \end{equation}
 where $\cos(\cdot,\cdot)$ denotes cosine similarity. Finally, our total loss for ID Conditional DDPM is as follows:
\begin{equation}
  \mathcal{L}_{\mathrm{total}} = \mathcal{L}_{\mathrm{id}} + \lambda \mathcal{L}_{\mathrm{noise}}.
  \label{eq:training_totalloss}
\end{equation}

The overall process for training ID Conditional DDPM is summarized in Alg.~\ref{alg:id_conditional_guided_diffusion}. Also, to verify the benefit of the ID Conditional DDPM, we conduct an ablation study, summarized in Table~\ref{tab:synth}. More examples of our ID Conditional DDPM are in the appendix.

\subsection{Facial Guidance} 
To control the facial attributes of generated images, we propose facial guidance that is applied during the diffusion process. One major advantage of using the diffusion model is that once the model is trained, it can control the image driven by the guidance during the sampling process. Thus we can obtain desired images without any re-training of the diffusion model. In order to utilize this advantage, we give facial guidance using external models, such as identity embedder \cite{deng2019arcface, wang2018cosface}, face parser \cite{yubilateral}, and gaze estimator \cite{Park2018ETRA} during the sampling process. Note that we can use any off-the-shelf facial model\cite{chang17fpn, chang17expnet}, and they can be adaptively selected according to the user's purpose.

\vspace{-10pt}
\paragraph{Identity Guidance.} In order to transmit the identity information of the source image, we employ the ID Conditional DDPM. However, this identity conditioning turned out to be insufficient, as the synthesized result image kept losing the source's identity. This is because when only we use the ID Conditional DDPM, the model focuses on other structural information, such as segmentation and gaze, so it fails to maintain source identity. Thus we give identity guidance to prevent loss of source identity in the denoising process. Specifically, we constrained the id vectors of the source and that of generated images to be located closer in the identity embedding space. Our facial identity guidance is formulated as follows: 
\begin{equation}
    \mathcal{G}_{\mathrm{id}} =   1-\cos\Big(\mathcal{D}_{\textit{I}}\big(\mathbf{x}_{\mathrm{src}}\big), \mathcal{D}_{\textit{I}}\big(\widehat{\mathbf{x}}_{0})\Big).
    \label{eq:identity_guidance}
\end{equation}

\begin{algorithm}[t]
    \footnotesize
    \caption{Diffusion-based Face Swapping with Facial Guidance}
    \label{alg:facial_guidance}
    \begin{algorithmic}
        \STATE \textbf{Input:} source image $\mathbf{x}_{\mathrm{src}}$, target image $\mathbf{x}_{\mathrm{targ}}$, target binary mask $M$, and masking threshold $\hat{T}$
        \STATE \textbf{Output:} swapped image $\mathbf{x}_{0}$ that reflects source identity while preserving id-irrelevant attributes in the target image $\mathbf{x}_{\mathrm{targ}}$
        \STATE $\mathbf{x}_T \sim \mathcal{N}(\mathbf{0}, \mathbf{I})$
        \STATE  ${\mathbf{v}}_{\mathrm{id}} \gets \mathcal{D}_{\textit{I}}({\mathbf{x}}_{\mathrm{src}})$
        \FORALL{$t$ from $T$ to $1$}
            \STATE $\mu, \sigma \gets \mu_{\theta}(\mathbf{x}_t), \sigma_{\theta}(\mathbf{x}_t)$
            \STATE $\widehat{\mathbf{x}}_0 \gets \frac{\mathbf{x}_t - \sqrt{1-\bar{\alpha}_t} \epsilon_{\theta}(\mathbf{x}_t, t, \textbf{v}_{\textrm{id}})}{\sqrt{\bar{\alpha}_t}} $
            
            \STATE $\mathcal{G}_{\mathrm{id}} \gets (1-\cos(\mathcal{D}_{\textit{I}}(\mathbf{x}_{\mathrm{src}}), \mathcal{D}_{\textit{I}}(\widehat{\mathbf{x}}_{0})))$
            
            \STATE $\mathcal{G}_{\mathrm{sem}} \gets ||\mathcal{D}_{\textit{F}}(\mathbf{x}_{\mathrm{targ}}) - \mathcal{D}_{\textit{F}}(\widehat{\mathbf{x}}_{0})||^2_2$

            \STATE $\mathcal{G}_{\mathrm{gaze}} \gets ||\mathcal{D}_{\textit{G}}(\mathbf{x}_{\mathrm{targ}}) - \mathcal{D}_{\textit{G}}(\widehat{\mathbf{x}}_{0})||^2_2$
            
            \STATE $\mathcal{G}_{\mathrm{facial}} \gets \lambda_{\mathrm{id}}\mathcal{G}_{\mathrm{id}} + \lambda_{\mathrm{sem}}\mathcal{G}_{\mathrm{sem}} + \lambda_{\mathrm{gaze}}\mathcal{G}_{\mathrm{gaze}}$
            
            \STATE $\widehat{\mathbf{x}}_{t-1} \sim \mathcal{N}(\mu - \sigma \nabla_{\textbf{x}_t}\mathcal{G}_{\mathrm{facial}}, \sigma)$
            \STATE $\mathbf{x}_{t-1,\mathrm{targ}} \sim \mathcal{N}(\sqrt{\bar{\alpha}_{t-1}} \mathbf{x}_{\mathrm{targ}}, (1-\bar{\alpha}_{t-1}) \mathbf{I})$
            
            \STATE $M_t \gets \min \{1, \frac{T-t}{\hat{T}}M\}$
            
            \STATE $\mathbf{x}_{t-1} \gets \widehat{\mathbf{x}}_{t-1} \odot M_t + \mathbf{x}_{t-1,\mathrm{targ}} \odot (1 - M_t)$
        \ENDFOR
        
        $\mathbf{x}_{\mathrm{swap}} \gets \mathbf{x}_0$
        \RETURN $\mathbf{x}_{\mathrm{swap}}$
    \end{algorithmic}
\end{algorithm}

\vspace{-15pt}
\paragraph{Semantic Guidance.} To explicitly match the facial features of the synthesized image to that of the target, we apply the face parsing model \cite{yubilateral}, which predicts pixel-wise labels for facial components (e.g., nose, eyebrows, and eyes). By leveraging a structural expert, we ensure that the generated image follows the facial structure of the target image.

As shown in Fig.~\ref{fig:main}, our DiffFace produces a frontal face guided by an ID expert at the beginning of the sampling process. Afterward, our semantic guidance induces the generated image to gradually follow the facial component of the target image. Finally, the generated image has a similar expression, pose, and shape as the target image by the provided semantic control. In this process, we composed a facial parsing map with essential labels like skin, eyes, and eyebrows, excluding non-facial components such as hair or glasses. We then compute the distance between two selected features. Our semantic guidance is formalized as follows:
\begin{equation}
    \mathcal{G}_{\mathrm{sem}} = \big|\big|\mathcal{D}_{\textit{F}}\big(\mathbf{x}_{\mathrm{targ}}\big) - \mathcal{D}_{\textit{F}}\big(\widehat{\mathbf{x}}_{0})\big|\big|^2_2 .
    \label{eq:sementic_guidance}
\end{equation}
\vspace{-20pt}
\paragraph{Gaze Guidance.} Gaze information plays a big role in conveying context and emotion. Considering that the face swap technique is heavily used in the film industry, preserving the gaze of the target is a critical issue. Nonetheless, previous models occasionally fail to preserve the target's gaze. This is because, without explicit gaze modeling, other terms are insufficient to guide gaze. Also, due to the diffusion stochastic process, synthesized results our model tend to have different gazes even if the same input, which makes it impossible to preserve the target's gaze. Thus, to solve these problems, we explicitly give guidance penalizing different gazes from the target image, using pretrained gaze estimating \cite{Park2018ETRA} module.

We first use the off-the-shelf facial landmark-detecting tool to obtain the coordinates of both eyes. Then we crop the eyes of the synthesized image and target image using the coordinates from the previous step. Next, we feed these crop eyes into a pretrained gaze estimating network \cite{Park2018ETRA} to obtain the gaze vectors. Finally, we calculate the distance between gaze vectors from the target image and the synthesized image and give it as gaze guidance. Our gaze guidance is formulated as follows:
\begin{equation}
    \mathcal{G}_{\mathrm{gaze}} = \big|\big|\mathcal{D}_{\textit{G}}\big(\mathbf{x}_{\mathrm{targ}}\big)-\mathcal{D}_{\textit{G}}\big(\widehat{\mathbf{x}}_{0})\big|\big|^2_2 .
    \label{eq:gaze_guidance}
\end{equation}

\vspace{-15pt}
\paragraph{Incorporationg Guidances.}
To guide the diffusion sampling process toward desired images, we incorporate the gradients from facial guidance modules. 
As shown in Eq.~\ref{eq:classifier_guidance}, we can induce facial expert models to behave like classifiers. In particular, given a diffusion model $\epsilon_\theta(\cdot)$ and pretrained facial expert models (e.g., $\mathcal{D}_{\textit{I}}$, $\mathcal{D}_{\textit{F}}$ and $\mathcal{D}_{\textit{G}}$), we can derive conditional sampling processes using these facial expert models. 
Our complete sampling procedure with the incorporated guidance is formulated as follows:
\begin{equation}
     \mathbf{x}_{t-1} \sim \mathcal{N}(\mu-\sigma\nabla_{\mathbf{x}_t}\mathcal{G}_\mathrm{facial}, \sigma),
\label{eq:guidance_facail}
\end{equation}
\vspace{-5pt}
where 
\begin{equation}
    \begin{split}
&\mu=\mu_\theta(\mathbf{x}_t,t,\mathbf{v}_\mathrm{id}),\quad
    \sigma=\sigma_\theta(\mathbf{x}_t,t,\mathbf{v}_\mathrm{id}), \\ &
    \mathcal{G}_\mathrm{facial} = \lambda_{\mathrm{id}}\mathcal{G}_{\mathrm{id}} + \lambda_{\mathrm{sem}}\mathcal{G}_{\mathrm{sem}}  + \lambda_{\mathrm{gaze}}\mathcal{G}_{\mathrm{gaze}},
    \end{split}
\label{eq:cost_volume}
\end{equation}    
 and $\mu_\theta(\cdot)$ is the predicted mean of $p_\theta(\mathbf{x}_t|\mathbf{x}_{t-1},\mathbf{v}_\mathrm{id})$. The overall process of facial guidance is summarized in Alg.~\ref{alg:facial_guidance}. Note that any additional facial expert models such as 3DMM~\cite{blanzMorphableModelSynthesis1999} and face pose estimation~\cite{ Ruiz_2018_CVPR_Workshops} can be added to our sampling procedure, which can further enhance the result to preserve the target attributes.

\subsection{Target-Preserving Blending}

Na\"ively using diffusion model in face swap task fails to preserve target background because every region is perturbed by noise. To preserve a background of the target with the desired identity, we utilize a face parser \cite{Yu-ECCV-BiSeNet-2018} which gives a semantic facial mask. We element-wise product the synthesized result with an obtained target hard mask $M$ to preserve the target's background while transferring the source facial identity. Similarly, previous methods ~\cite{avrahami2022blended, lugmayr2022repaint, meng2021sdedit} use a strategy of blending two noisy images with the user-specified binary mask. However, when blending the image in the face swap task with a hard mask, the structural attributes of the target image are removed by noise due to the strong mask intensity. 

To achieve our goals, we propose a target-preserving blending method that alters the mask intensity to better preserve structural attributes of target. Target-preserving blending is to gradually increase the mask intensity from zero to one, according to the time of the diffusion process $T$. By adjusting the starting point where the intensity of the mask becomes one, we can adaptively maintain the structure of the target image. In this manner, we effectively control the trade-off relationship between identity and structure attributes in the face swap task, which will be shown in Sec.~\ref{sec:ablation_study}.

Fig.~\ref{fig:arch} shows an overview of target-preserving blending. $M$ denotes the hard mask obtained by the face parser, and $M_{t}$ denotes the soft mask whose intensity is strengthened over time. $\hat{T}$ denotes the starting point where the mask intensity becomes one as a hard mask. The soft mask $M_{t}$ is obtained:
\vspace{-5px}
\begin{equation}
    M_{t} = \min (1, \frac{T-t}{\hat{T}}M),
    \label{eq:target_preserving_mask}
\end{equation}
We blend the intermediate prediction in the reverse process and target image using the mask $M_{t}$. Specifically, we redefine $\mathbf{x}_{t-1}$ in Eq.~\ref{eq:guidance_facail} to $\widehat{\mathbf{x}}_{t-1}$ as follows:
\begin{equation}
    \widehat{\mathbf{x}}_{t-1} \sim \mathcal{N}(\mu-\sigma\nabla_{\mathbf{x}_t}\mathcal{G}_\mathrm{facial}, \sigma).
\end{equation}
Then, to match the noise level between the target image $\mathbf{x}_\mathrm{targ}$ and the intermediate prediction $ \widehat{\mathbf{x}}_{t-1}$, we derive a noisy target image $\mathbf{x}_{t-1,\mathrm{targ}}$. Finally, we blend the intermediate prediction $ \widehat{\mathbf{x}}_{t-1}$ and the noisy target image $\mathbf{x}_{t-1,\mathrm{targ}}$ as
\begin{equation}
    \mathbf{x}_{t-1} = \widehat{\mathbf{x}}_{t-1} \odot M_{t} + \mathbf{x}_{t-1,\mathrm{targ}} \odot (1 - M_{t}).
     \label{eq:target_preserving_masking}
\end{equation}
We define $\mathbf{x}_{\mathrm{swap}}$ the result of Eq.~\ref{eq:target_preserving_masking} when $t=1$. Please refer to Alg. \ref{alg:facial_guidance} for more details.
\section{Experiment}
\label{sec:experiment}

\subsection{Implementation Details}

\paragraph{Training.} Our ID Conditional DDPM follows the architecture of U-Net~\cite{https://doi.org/10.48550/arxiv.1505.04597} based on Wide ResNet~\cite{zagoruyko2016wide}. We train our ID Conditional DDPM on FFHQ\cite{karras2019style} dataset, which consists of 70k aligned face images with resolution 256 × 256. We set $\lambda=0.5$ at training. We implement our network using PyTorch~\cite{paszke2019pytorch}, and AdamW optimizer~\cite{https://doi.org/10.48550/arxiv.1711.05101} is used for training, where the learning rate is set to 0.0001. We train the model 700k steps with batch size 48 for approximately 10 days on 8 NVIDIA A100 PCIe 80GB GPUs. 

\vspace{-10pt}
\paragraph{Sampling.} In the sampling process, we use extending augmentations used in \cite{avrahami2022blended} to prevent adversarial results. The number of extending augmentation is set to 8. We chose T = 75 for the number of diffusion steps so that the model could sufficiently alter the target image. We set weights in facial guidance $\lambda_{\mathrm{id}}$, $\lambda_{\mathrm{sem}}$, and $\lambda_{\mathrm{gaze}}$ to 2000, 150, and 200, respectively. 


\begin{figure*}[ht]
    \centering
    \includegraphics[width=1\textwidth]{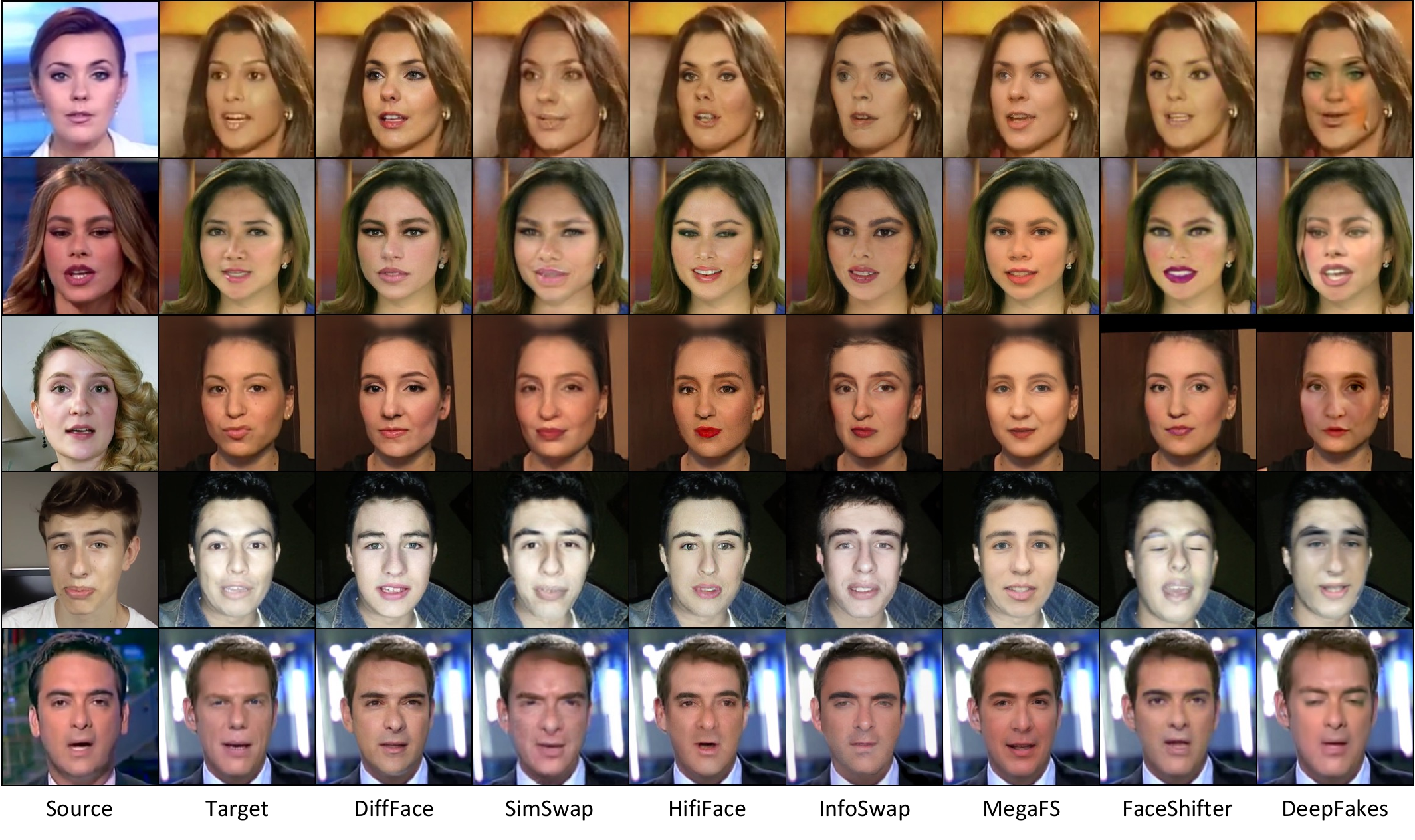}
    \vspace{-25pt}
    \caption{
    \textbf{Qualitative comparison of face swap results with other various models.} The results of our model better reflects the source identity while successfully removing target identity-related attributes. Refer Sec.~\ref{sec:comparison-with-baselines} for more analysis. 
    }
    \label{fig:compare}
    \vspace{-10pt}
\end{figure*}

\subsection{Evaluation Protocol}
\paragraph{Evaluation Dataset.} We evaluate DiffFace on FaceForensics++ (FF++) dataset \cite{rosslerFaceForensicsLearningDetect2019}. FaceForensics++ dataset is a standard dataset for evaluating face swap methods. All frames are uniformly sampled from the 1000 original videos and generated from various face swap models.

\vspace{-10pt}
\paragraph{Compared Models.} We compare our model with state-of-the-art face swap methods. Because this model is the first to adapt diffusion models to face swap tasks, we evaluate our model with traditional GAN-based methods. We use SimSwap \cite{chen2020simswap}, HifiFace \cite{wang2021hififace},  InfoSwap \cite{gao2021information},  MegaFS \cite{zhu2021one}, FaceShifter \cite{liFaceShifterHighFidelity2019}, and Deepfakes \cite{DeepFakesHttpsGithub2021}.

\vspace{-10pt}
\paragraph{Quantitative Evaluations.} We only sample a single image from our model for objective comparison, despite the fact that our model can output diverse images. We quantitatively evaluate models using ID cosine similarity, expression, pose, and shape. For identity metric, we use ArcFace and CosFace embedder to compute the embedding distance between $\mathbf{x}_{\mathrm{swap}}$ and $\mathbf{x}_{\mathrm{src}}$. For a fair comparison, we use ArcFace \cite{deng2019arcface} and CosFace \cite{wang2018cosface} to measure models trained with CosFace and ArcFace, respectively. Also, we apply the relative distances metric which was proposed in SmoothSwap \cite{kimSmoothSwapSimpleEnhancement2021} in order to measure not only how close $\mathbf{x}_{\mathrm{swap}}$ and $\mathbf{x}_{\mathrm{src}}$ is, but also how far $\mathbf{x}_{\mathrm{swap}}$ and $\mathbf{x}_{\mathrm{targ}}$ is. The relative distance metric is formalized in Eq.~\ref{eq:relative}. To evaluate the expression, pose and shape, we use pretrained network\cite{RingNet:CVPR:2019} to obtain coefficients of 3D face model\cite{FLAME:SiggraphAsia2017}, then compute the distance as follows:

\begin{equation}
     \mathcal{D}-\mathbf{R}:= \frac{\mathcal{D}(\mathbf{x}_\mathrm{swap},\mathbf{x}_\mathrm{src})}{\mathcal{D}(\mathbf{x}_\mathrm{swap},\mathbf{x}_\mathrm{src})+\mathcal{D}(\mathbf{x}_\mathrm{swap},\mathbf{x}_\mathrm{targ})},
\label{eq:relative}
\end{equation}
where $\mathcal{D}$ can be any distance metric, e.g. cosine distance.
\tabPerformance
\smallskip

\subsection{Comparison with Baselines}
\label{sec:comparison-with-baselines}
Fig.~\ref{fig:compare} shows that our DiffFace outperforms other models in terms of changing identity-related attributes. For example, in the first and fourth rows, we notice our result reflects more vivid lips and eyes, while other results models tend to have eyes and lip colors from target images. This shows that our model more effectively transfers identity-related attributes than other models. 

Table~\ref{tab:comparison} shows the same tendency as the qualitative result shown above. As shown in the first four columns, our model achieves the highest identity score. Also, the ability to remove the target's identity is superior to any other model. We speculate that these results are caused by the unique property of the diffusion process. Specifically, existing GAN-based models could not remove specific regions, while diffusion models can remove target information with random noise.

\tabAblation

Although our approach yields limited performance on expression, pose, and shape score, one thing to note here is that we can manipulate guidance in the sampling process in order to compensate the problem. Specifically, we can adjust the reflectivity of the target image in the reverse process by controlling $\hat{T}$, which is the starting point where the intensity of the mask becomes one.  In summary, we can dynamically alter our model to satisfy different purposes. Please refer to Sec.~\ref{sec:ablation_study} for more details. 

\begin{figure*}[ht]
    \centering
        \begin{subfigure}{0.3486976744186046\linewidth}
        \includegraphics[width=\textwidth]{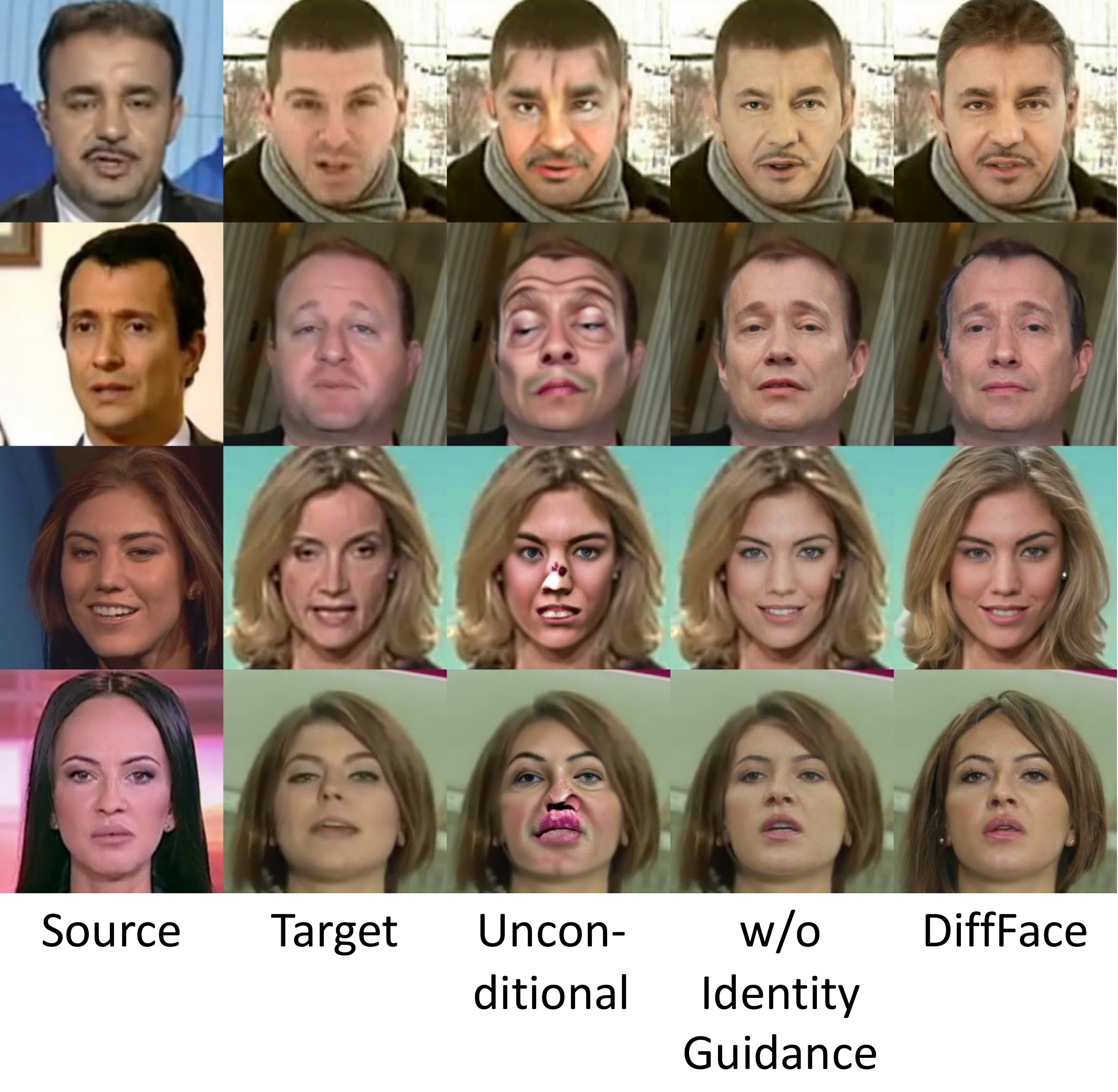}
        \caption{Ablation for ID Conditional DDPM}
        \label{fig:short-a}
      \end{subfigure}
      \hfill
      \begin{subfigure}{0.3504069767441861\linewidth}
       \includegraphics[width=\textwidth]{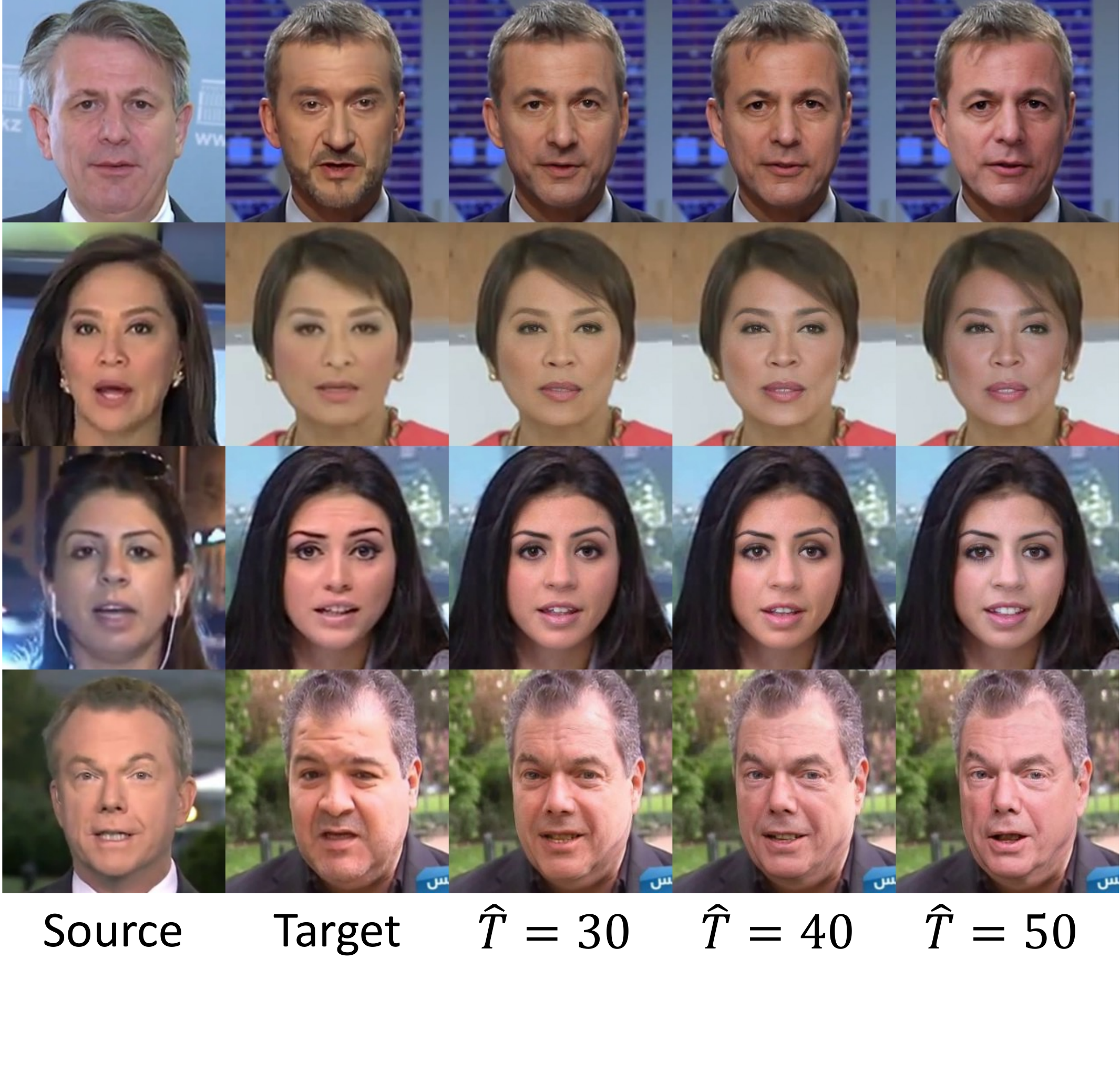}
        \caption{Ablation for target preserving blending}
        \label{fig:short-b}
      \end{subfigure}
      \hfill
      \begin{subfigure}{0.2808953488372093\linewidth}
       \includegraphics[width=\textwidth]{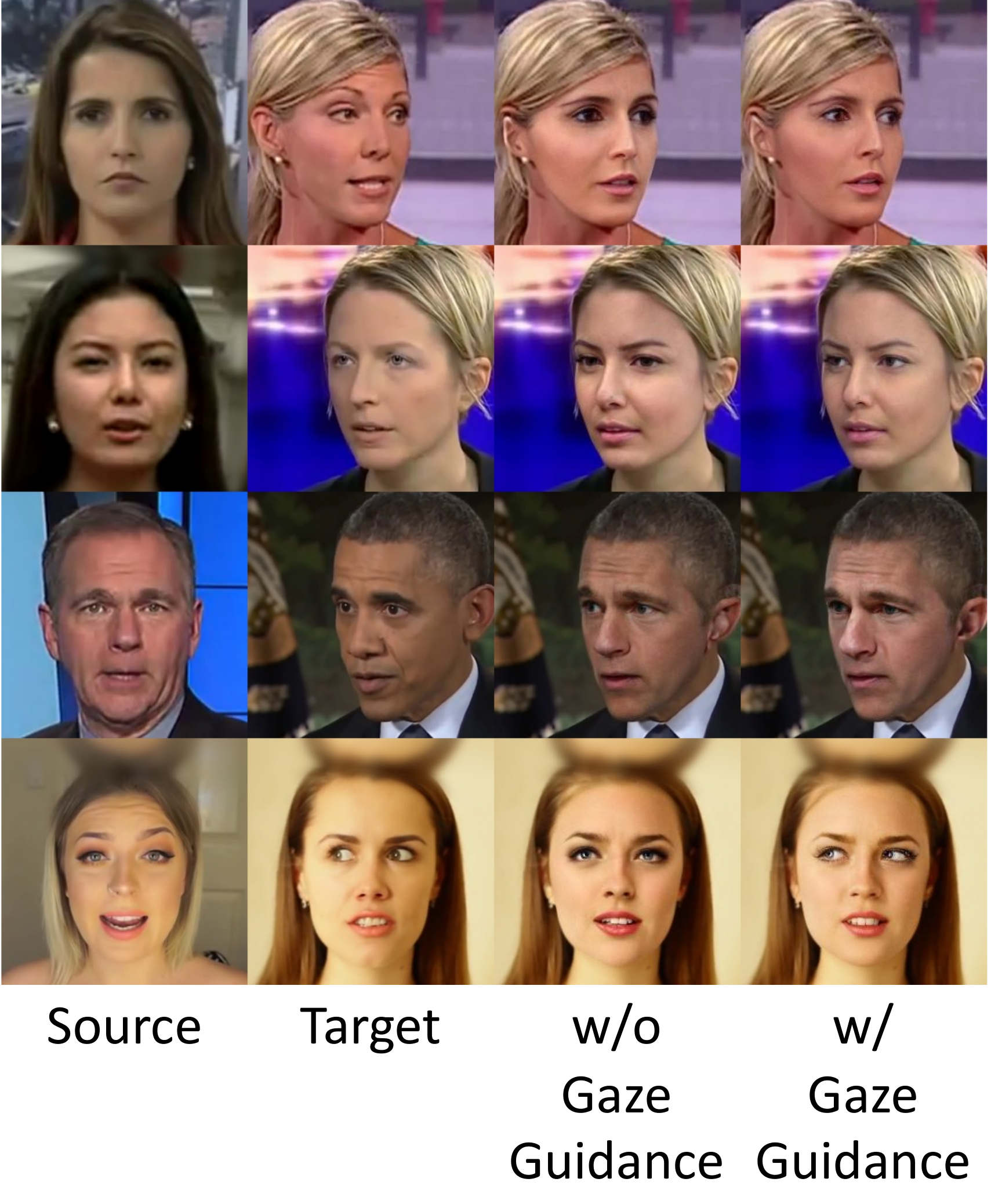}
        \caption{Ablation for gaze guidance}
        \label{fig:short-b}
      \end{subfigure}
      \vspace{-5pt}
    \caption{\textbf{Qualitative results of ablations:} (a) ID Conditional DDPM, (b) target preserving blending, and (c) gaze guidance. Our method can be easily controlled without any additional training. We can obtain various face swapped results by controlling the facial module. 
    }
    \label{fig:ablation}
    \vspace{-10pt}
\end{figure*}

\vspace{-2pt}
\subsection{Ablation Study and Analysis}
\label{sec:ablation_study}

\vspace{-5pt}
\paragraph{Identity Guidance.}
\textbf{(I)} and \textbf{(II)} in Table~\ref{tab:synth} show a quantitative ablation study on ID Conditional DDPM and identity guidance. Our model results in the highest identity similarity score when ID Conditional DDPM and identity guidance are present. 
The same trend can be found in Fig.~\ref{fig:ablation}~(a), which shows a qualitative ablation study on ID Conditional DDPM and identity guidance. It is noticeable that results without ID Conditional DDPM show visible artifacts, and results without identity guidance show poor identity transfer compared to the full model. 

\begin{figure}[]
    \begin{center}
    \includegraphics[width=0.85\linewidth]{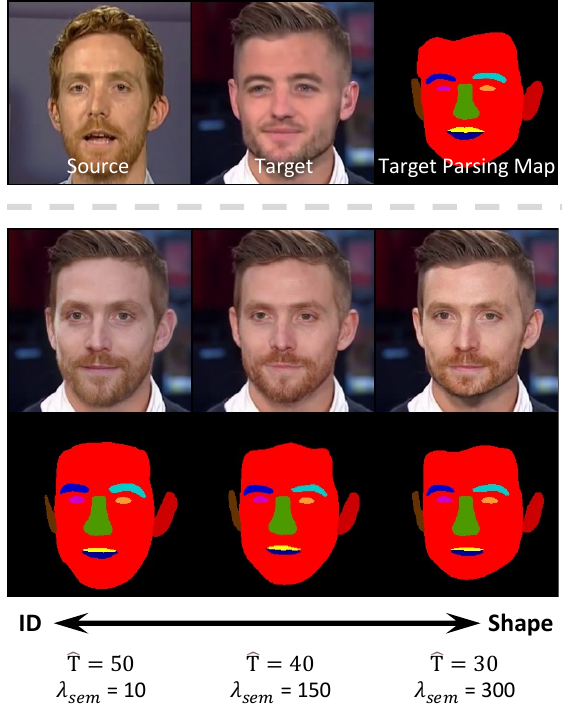}
    \caption{\textbf{Trade-off between ID and shape.} The results show how our model can be altered to prioritize either shape or id. The second and third row represents the synthesized image and corresponding face parsing map.}
    \vspace{-20pt}
    \label{fig:ablation_tradeoff}
    \end{center}
\end{figure}

\vspace{-10pt}
\paragraph{Target Preserving Blending.}
In this ablation study, we adjust $\hat{T}$, the starting point where the mask becomes one as a hard mask, to investigate the effect of target preserving blending, as shown in \textbf{(III)} to \textbf{(VIII)} of Table~\ref{tab:synth}. As $\hat{T}$ increases, the id similarity score also increases because the model gains more control over the image and thus manipulates the image close to the source image. At the same time, expression, pose, and shape deteriorate because the image gets manipulated more than before. The same tendency is also shown in Fig.~\ref{fig:ablation}~(b). As $\hat{T}$ decreases, the synthesized image tends to move toward the target image, in terms of expression, pose and shape. Also, we can observe that the skin tone of the synthesized image approaches the target. In summary, we can choose our model to balance between preserving the target attributes and injecting source identity without additional training.   

\vspace{-10pt}
\paragraph{Gaze Guidance.}
Fig.~\ref{fig:ablation}~(c) illustrates our qualitative ablation study on gaze guidance. Recall that the synthesized image's gaze must follow the target image. We can find that with gaze guidance, the synthesized image has the same gaze as the target image. Thus we employ gaze guidance to fulfill gaze information which was not controllable using the identity or semantic guidance.

\vspace{-10pt}
\paragraph{Trade-off Control on ID and Shape.}
Fig.~\ref{fig:ablation_tradeoff} shows how our model can be altered to prioritize either ID or shape. We can induce our model to show active shape change by incrementing $\hat{T}$ and decrementing $\lambda_{\mathrm{sem}}$. Conversely, we can force the model to focus on preserving target structure by decrementing $\hat{T}$ and incrementing $\lambda_{\mathrm{sem}}$. As a result, we can see that the shape of the chin on the left is gradually getting rounder as it goes to the right. In this way, we can easily have controllability without the need to retrain the model in the face swap task.

\section{Conclusion}
\label{sec:conclusion}


In this work, we propose \textit{DiffFace}, a diffusion-based framework aiming for controllable and high-fidelity subject-agnostic face swapping. Based on the trained our ID Conditional DDPM, the facial guidance from the pretrained facial expert models make the model to faithfully transfer the source identity while preserving target attributes during the sampling process. The target preserving blending helps our framework to keep the attributes of the target face from noise while transferring the source facial identity. Moreover, our framework can flexibly apply various facial guidance and adaptively control the ID-attribute tradeoff to achieve better results without additional training. Extensive experiments demonstrate that the proposed framework significantly outperforms previous face swapping methods.

{\small
\bibliographystyle{ieee_fullname}
\bibliography{references}
}

\clearpage


\crefname{section}{Sec.}{Secs.}
\Crefname{section}{Section}{Sections}
\Crefname{table}{Table}{Tables}
\crefname{table}{Tab.}{Tabs.}

\maketitle

\appendix

{\LARGE \textbf{Appendix}}

\section{Architecture Details}
\subsection{Identity Embedding Model}
We employ a pretrained ArcFace~\cite{deng2019arcface} and CosFace~\cite{an2022pfc} identity embedding models to extract the identity vector of the source face. ArcFace and CosFace models are based on ResNet-101~\cite{he2016deep} and the identity vector becomes an unit-length ($\Vert v_{id} \Vert=1$) by the \textit{UnitNorm} in Fig.~\ref{supple-fig:IDCondDDPM}.

\subsection{ID Conditional DDPM}     \label{supple:IDCondDDPM}

\begin{figure}[!ht]
    \centering
    \includegraphics[width=0.8\columnwidth]{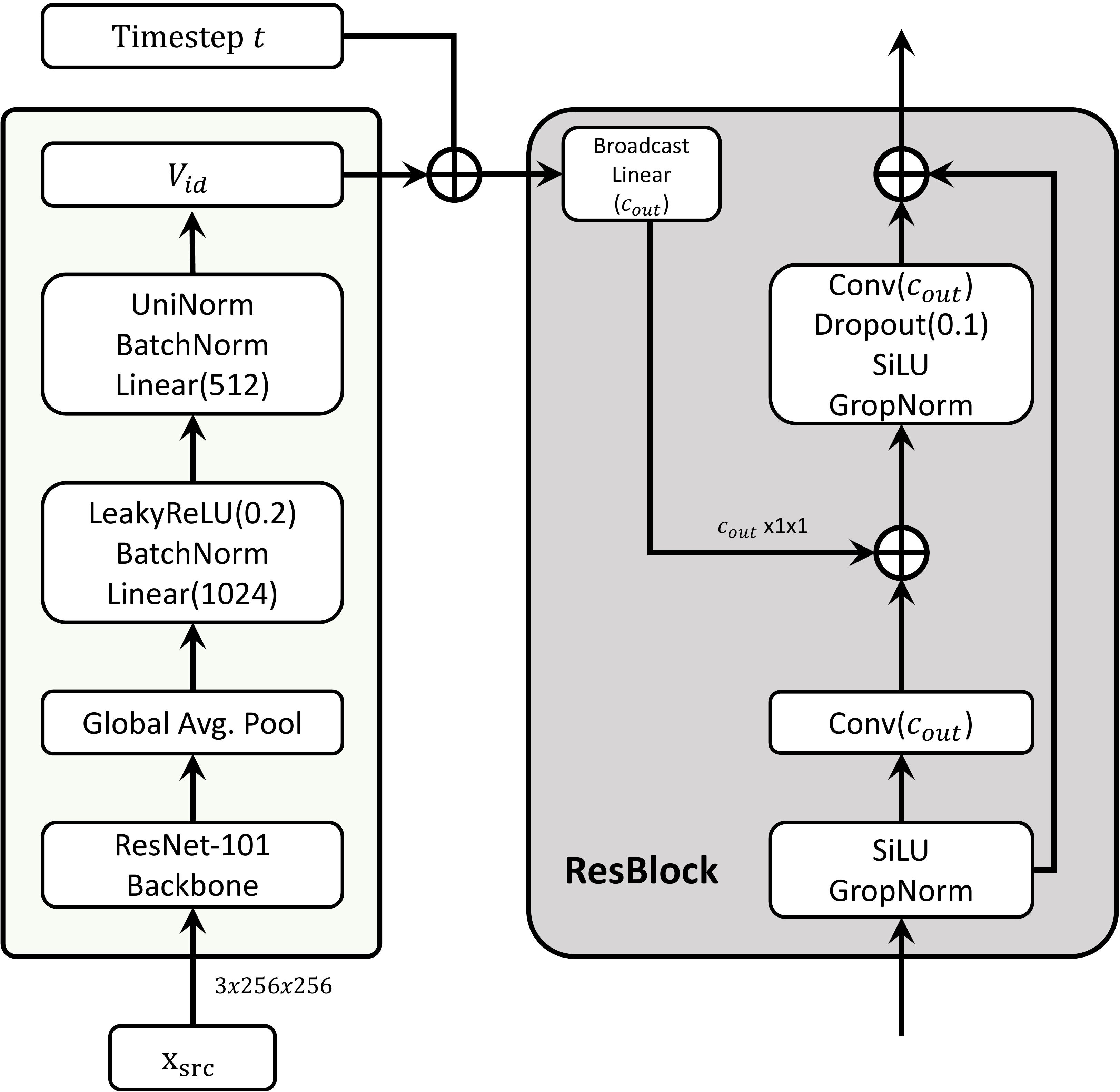}
    \caption{The residual block structure of ID Conditional DDPM. The identity vector $v_{id}$ is embedded into each Resblocks of U-Net with the diffusion timestep $t$. We denote $\oplus$ the summation operation and feature dimensions are written in the order of (channels $\times$ height $\times$ width)}
    \vspace{-10pt}
    \label{supple-fig:IDCondDDPM}
\end{figure}

\begin{figure*}
    \centering
    \includegraphics[width=0.9\textwidth]{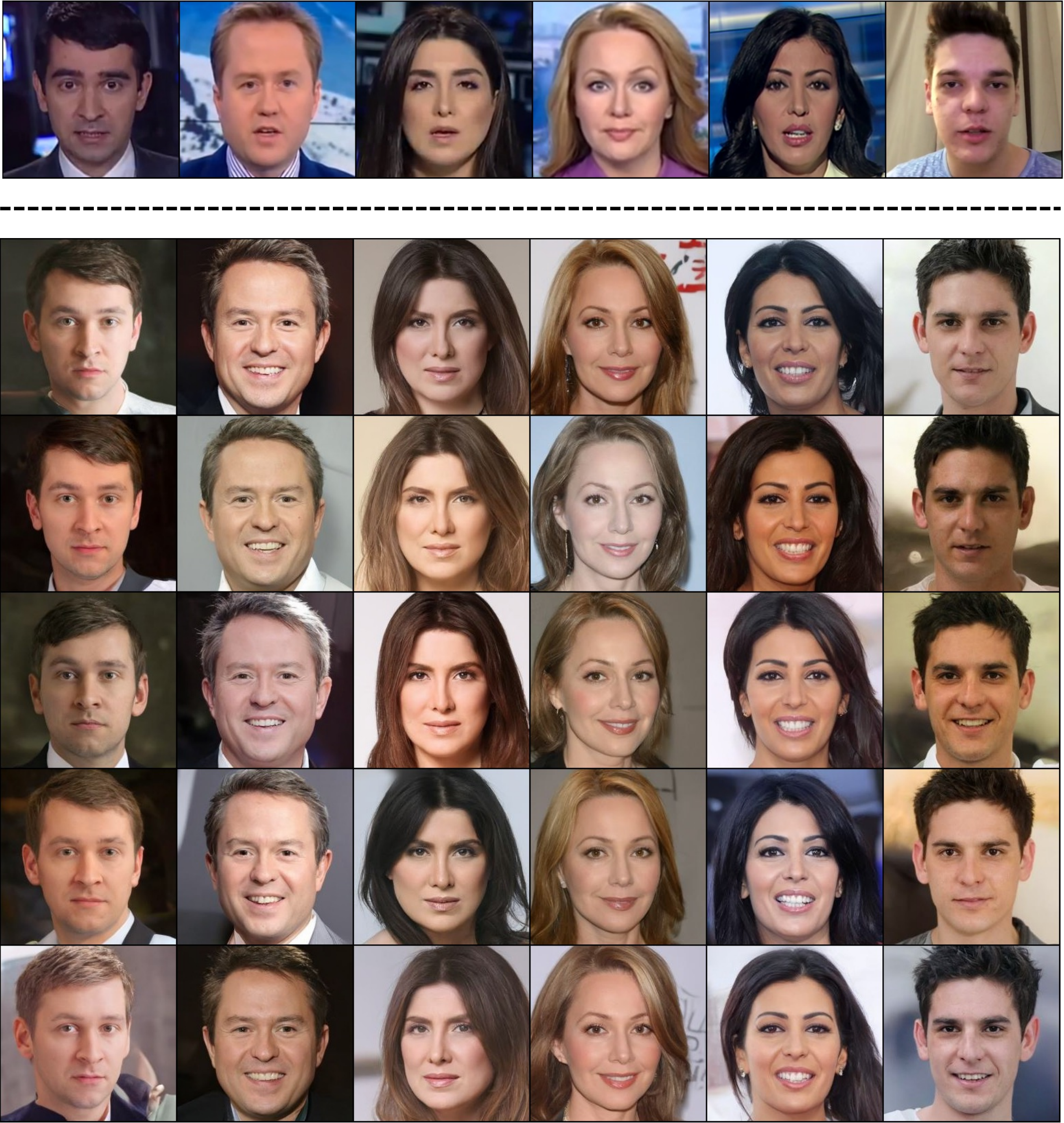}
    \caption{Image samples from ID Conditional DDPM (Sec.~\ref{supple:IDCondDDPM}). The first row is the source faces which are provided to the identity embedder~\cite{deng2019arcface,an2022pfc}. The last five rows are samples generated by ID Conditional DDPM.
    }
    \label{supple-fig:id_sample}
\end{figure*}

Our diffusion model follows the architecture of guided diffusion model\cite{dhariwal2021diffusion}, except for conditioning the identity. In order to sample results with the desired identity, we inject the identity vector obtained from the identity embedder~\cite{deng2019arcface, an2022pfc} into the residual blocks of U-Net. As a result, we can condition our diffusion model and sample images with specific identities. The detailed structure of ID Conditional DDPM is depicted in Fig.~\ref{supple-fig:IDCondDDPM}. Also, we show some examples of ID Conditional DDPM in Fig.~\ref{supple-fig:id_sample}.

\section{Sampling Details}
\subsection{Face Parser} \label{supple-sec:gen}
In every timestep of the sampling process, we use pretrained face parsing network\cite{Yu-ECCV-BiSeNet-2018} to give facial guidance. The pretrained face parsing network outputs 19 classes, including all facial components and accessories. We construct a facial mask using 11 classes related to the face swap task (e.g., skin, nose, and eyes) and exclude non-facial components (e.g., hair, a hat, and cloth).


\subsection{Gaze Estimator} \label{supple-sec:gen}
Similar to face parser, we give gaze guidance at every time step during the sampling process. We first use facial landmark detecting tool~\cite{bulat2017far} to obtain coordinates of both eyes and then crop eye images. The eye images are automatically resized to 96x160. Then the images are fed into the pretrained gaze estimator~\cite{Park2018ETRA} to obtain gaze-relevant features.


\section{Additional Results from DiffFace}

\subsection{Target-Preserving Blending}
Fig.~\ref{supple-fig:progress} shows the change of the noisy image $x_t$ and facial mask $M_t$ during the diffusion process. Our facial mask intensity increases linearly until the masking threshold $\hat{T}$. By altering the mask's intensity, which determines the reflectivity of the target image, we can preserve the fine details of the facial attributes (e.g., expression, pose and skin tone).

\subsection{Trade-off between ID and Shape}
\begin{figure*}
  \centering
  \begin{subfigure}[b]{\textwidth}\centering
  \includegraphics[width=0.6\textwidth]{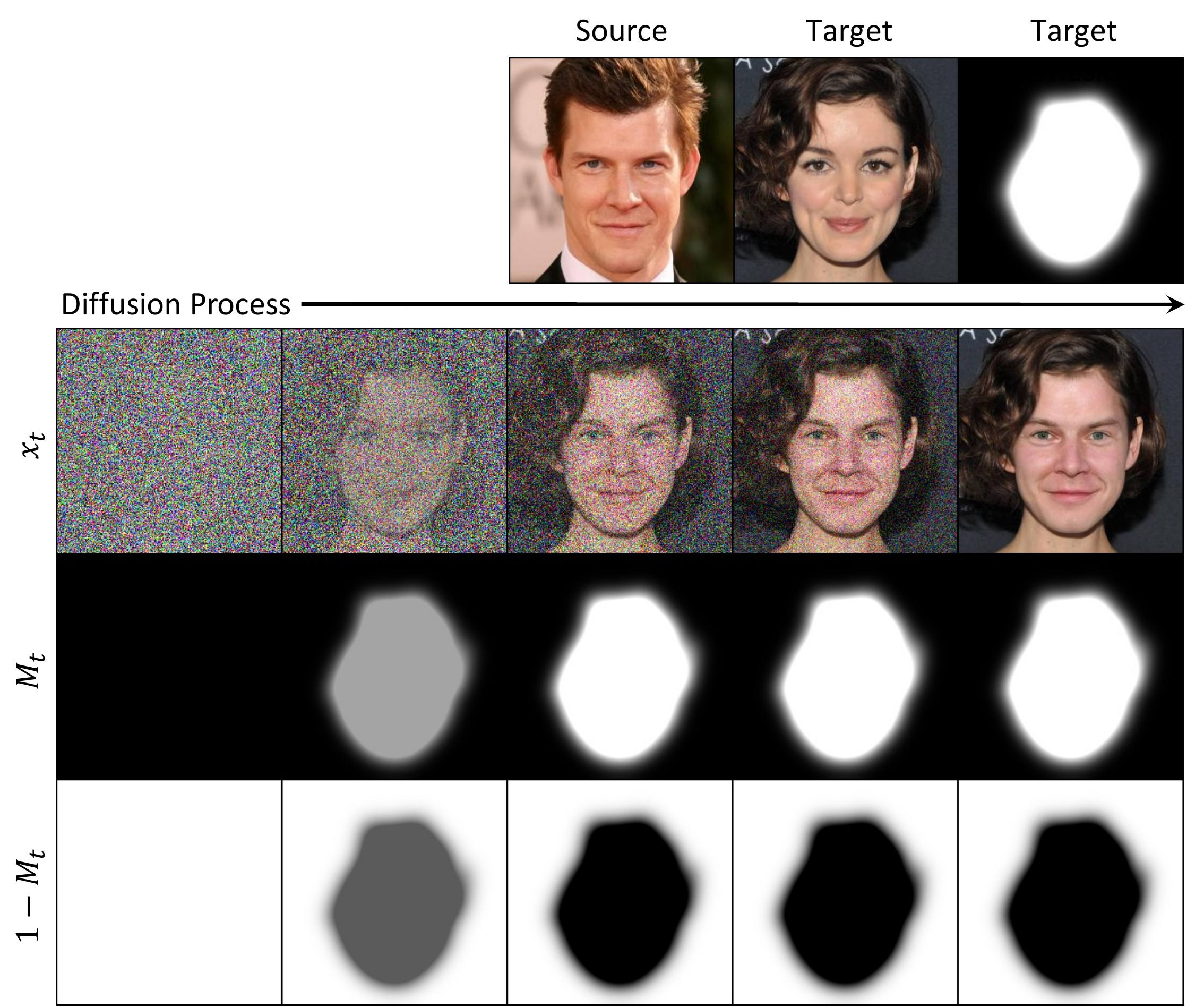}
    \caption{Visualization result of a noisy image $x_t$, facial mask $M_t$ and background mask $1-M_{t}$ for the reverse process.}
    \label{supple-fig:progress}
  \end{subfigure}
  
  \vspace{20pt}
  
  \begin{subfigure}[b]{\textwidth}\centering
    \includegraphics[width=0.6\textwidth]{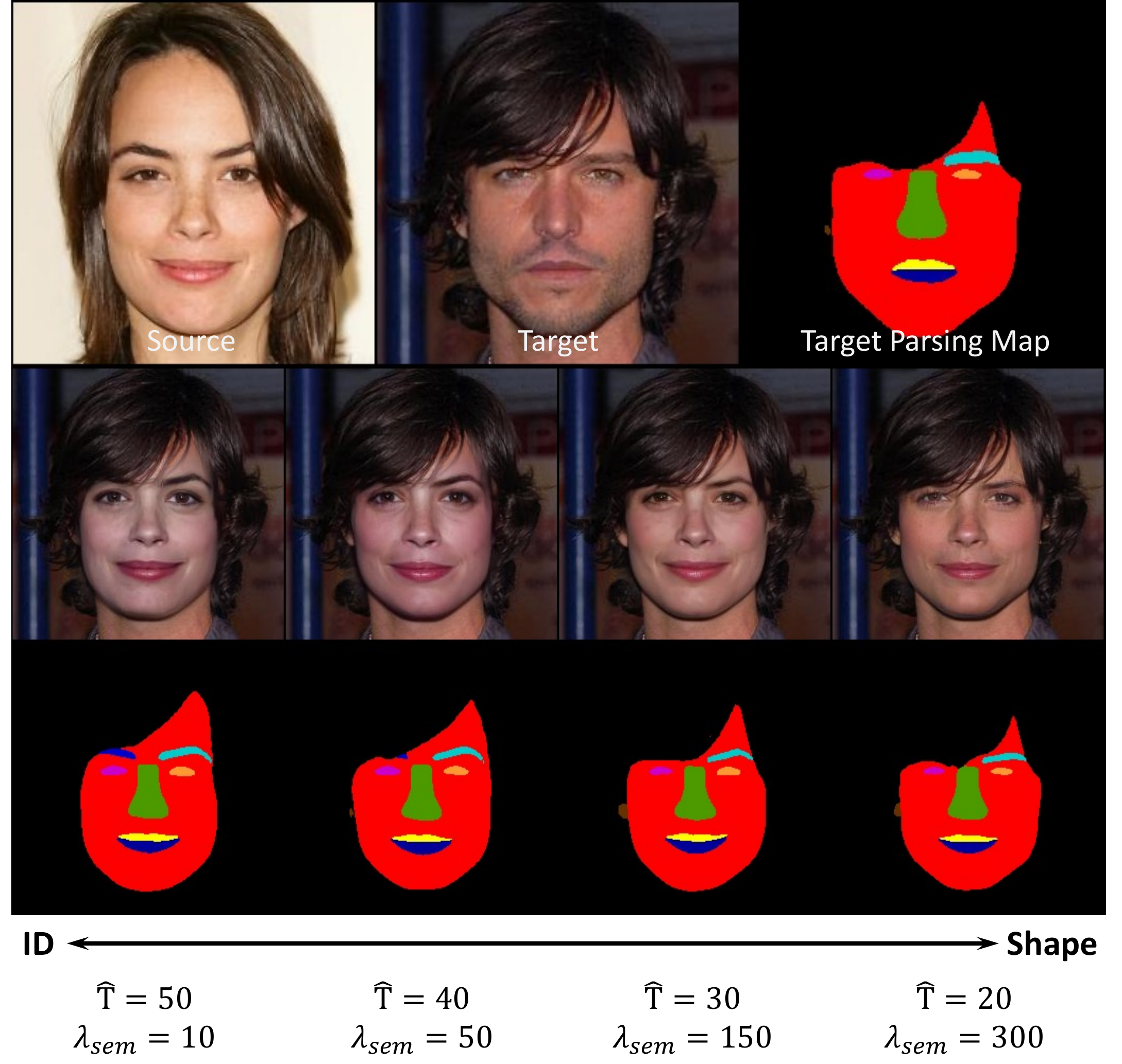}
    \caption{Visualization result for the trade-off between ID and Shape by controlling $\hat{T}$ and $\lambda_{\mathrm{sem}}$}
    \label{supple-fig:trade-off}
  \end{subfigure}
    \caption{More ablation results for Target-Preserving Blending and Trade-off between ID and Shape}
  \label{supple-fig:ffpp}
\end{figure*}

Unlike GAN-based face swapping method that produces a deterministic output image, our method can control the face shape with the facial guidance $\mathcal{G}_\mathrm{facial}$ from $\lambda_{\mathrm{sem}}$ and masking threshold $\hat{T}$. Fig.~\ref{supple-fig:trade-off} shows the result of emphasizing ID or face shape. For example, the sample in the first column of the third row shows a round chin shape, while the sample in the fourth column of the third row shows a sharp chin shape. Also, we can observe the skin tone changes in various ways according to the time threshold $\hat{T}$.
 
\begin{figure}[!ht]
    \centering
    \includegraphics[width=\columnwidth]{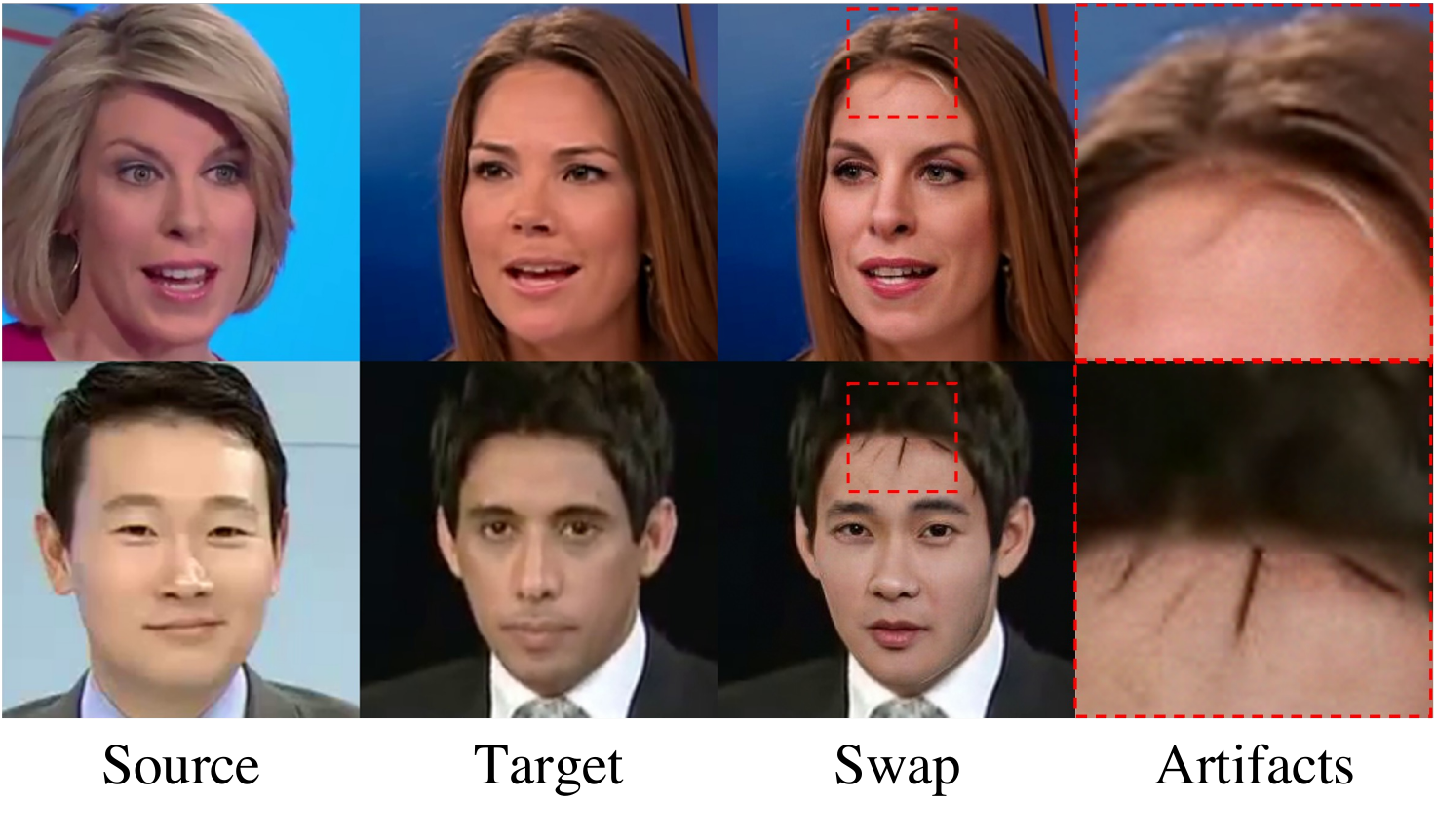}
    \vspace{-20pt}
    \caption{Some failure cases of DiffFace on the FaceForensics++ dataset \cite{rosslerFaceForensicsLearningDetect2019}.}
    \label{supple-fig:limit}
\end{figure} 

\subsection{Comparison and Out-Of-Domain Results}
\begin{figure*}
    \centering
    \includegraphics[width=\textwidth]{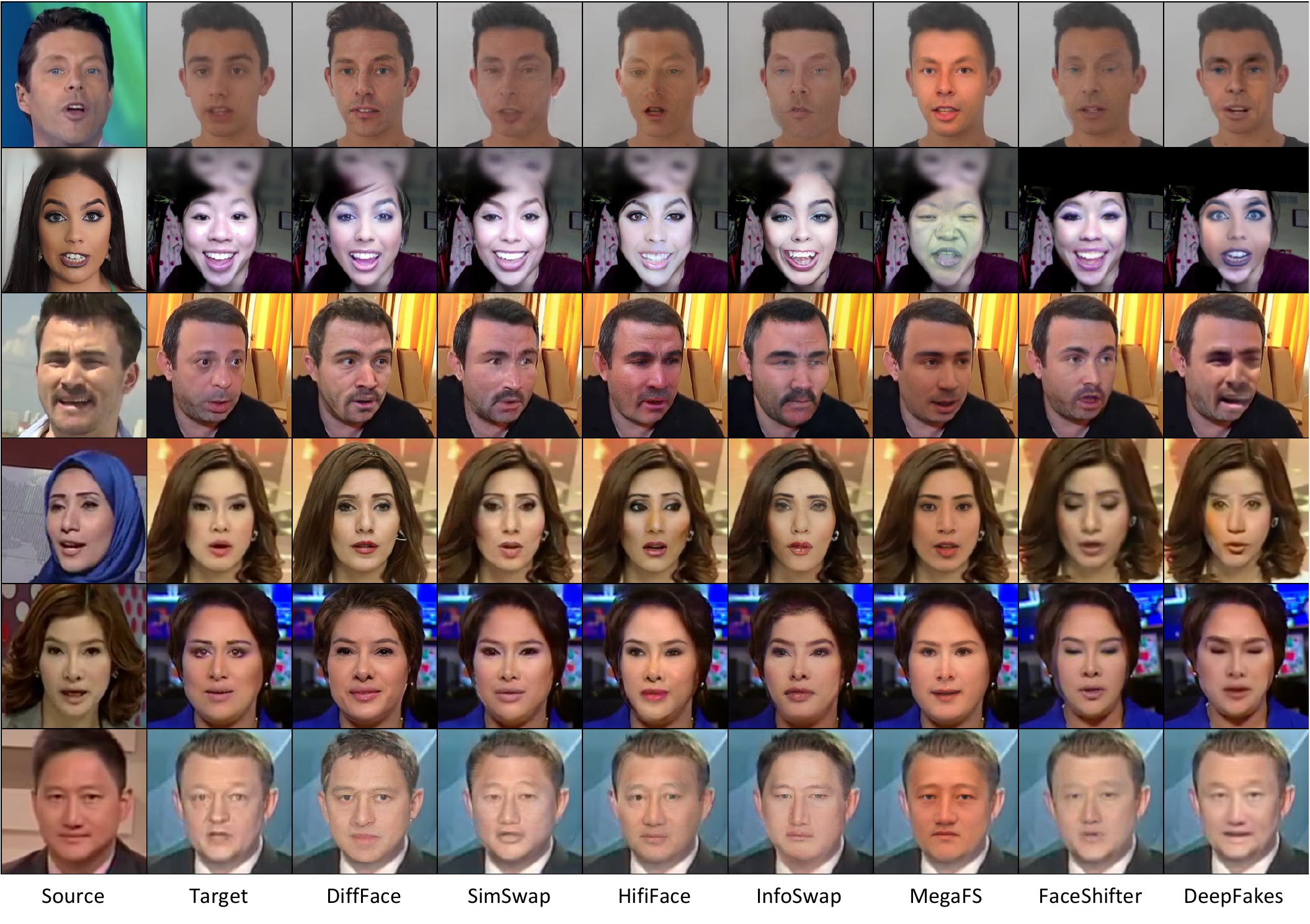}
    \caption{Additional comparison results of diverse models on the FaceForensics++ dataset \cite{rosslerFaceForensicsLearningDetect2019}.}
    \label{supple-fig:comparison}
\end{figure*}

\begin{figure*}
    \centering
    \includegraphics[width=\textwidth]{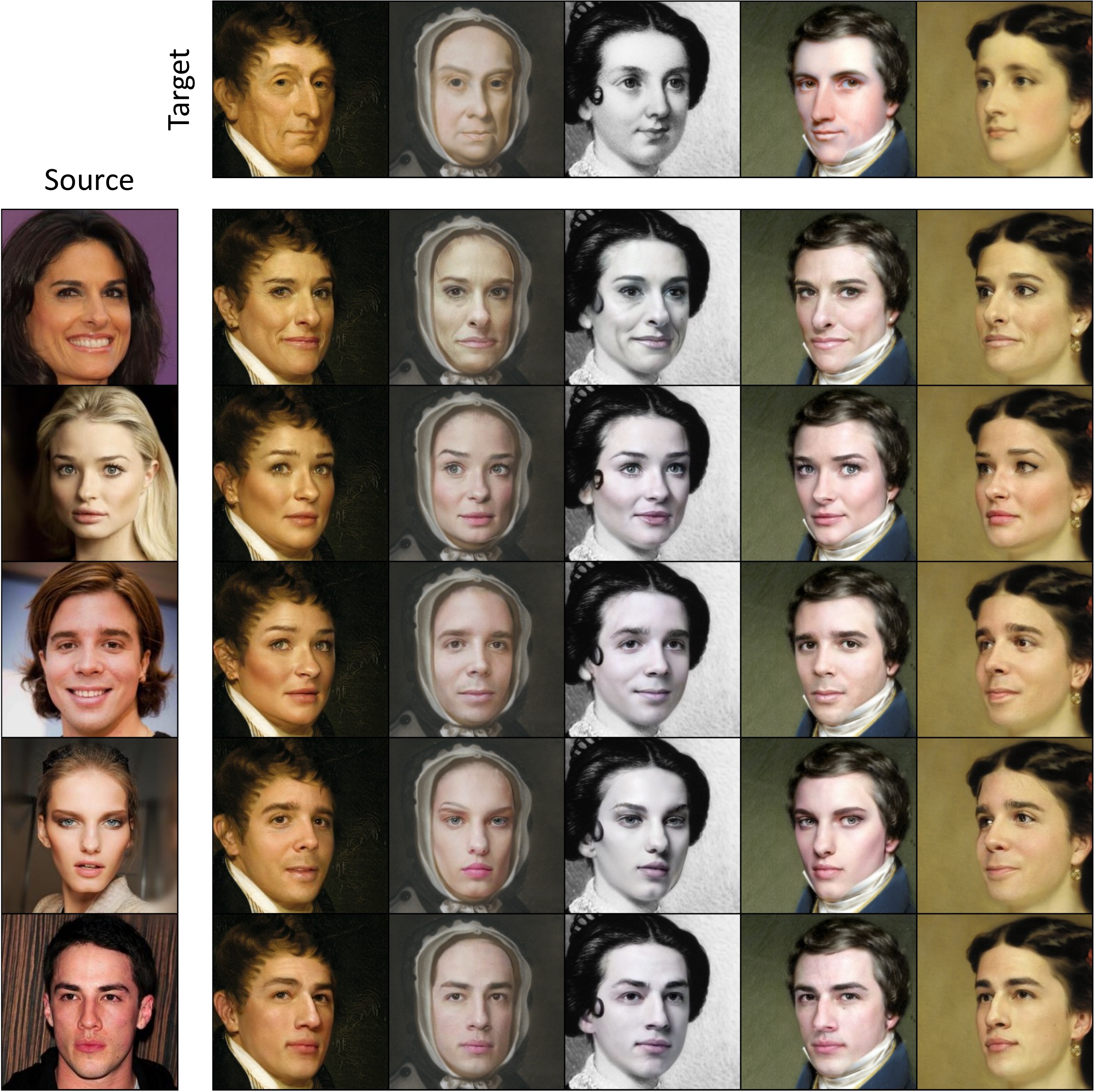}
    \caption{Out-of-domain face swapping results generated by our DiffFace on the MetFaces dataset~\cite{rosslerFaceForensicsLearningDetect2019}.}
    \label{supple-fig:Metface}
\end{figure*}

\begin{figure*}
    \centering
    \includegraphics[width=\textwidth]{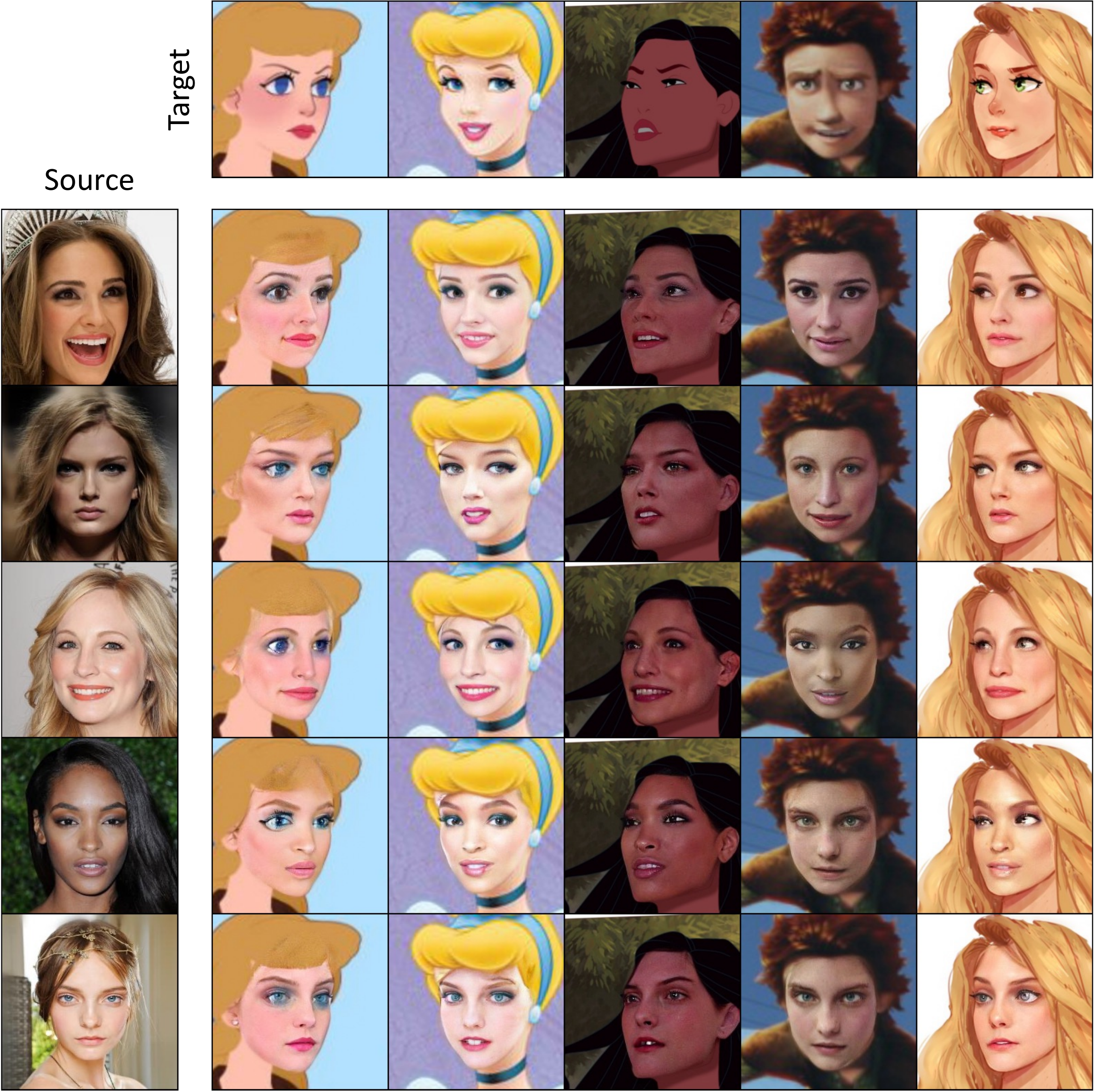}
    \caption{Out-of-domain face swapping results generated by our DiffFace on the Disney Face dataset~\cite{pinkney2020resolution}.}
    \label{supple-fig:cartoon}
\end{figure*}

We provide additional collections of swapped-image samples based on our DiffFace model. Fig.~\ref{supple-fig:comparison} show the extra face swapping results of various models on the FaceForensic++ dataset~\cite{rosslerFaceForensicsLearningDetect2019}. Fig.~\ref{supple-fig:Metface} and Fig.~\ref{supple-fig:cartoon} show the results of out-of-domain datasets, where oil portrait paintings (Metfaces dataset~\cite{karras2020training}) and cartoon faces (Disney Face dataset~\cite{pinkney2020resolution}) are used. Although our model is not trained on any of these images, the results reflects the characteristics of each domain with shape changing.

\section{Limitations}
\label{sec:limitation}

Denoising diffusion probabilistic models have shown remarkable generative performance in various computer vision tasks. However, since the diffusion model generates images through sequential stochastic transitions, once an artifact (i.e., wrinkles, hair segments, and glasses) occurs, it is easy to maintain it (Fig. \ref{supple-fig:limit}). For this reason, our model sometimes shows unintended artifacts in the output. Therefore, research to correcting transitions with artifacts seems to be necessary.


\end{document}